\documentclass[sigconf,screen,authorversion]{acmart}
\AtBeginDocument{%
  \providecommand\BibTeX{{%
    \normalfont B\kern-0.5em{\scshape i\kern-0.25em b}\kern-0.8em\TeX}}}

\setcopyright{iw3c2w3}
\copyrightyear{2021}
\acmYear{2021}
\setcopyright{iw3c2w3}
\acmConference[WWW '21]{Proceedings of the Web Conference 2021}{April 19--23, 2021}{Ljubljana, Slovenia}
\acmBooktitle{Proceedings of the Web Conference 2021 (WWW '21), April 19--23, 2021, Ljubljana, Slovenia}
\acmPrice{}
\acmDOI{10.1145/3442381.3450034}
\acmISBN{978-1-4503-8312-7/21/04}

\usepackage{amsmath}
\usepackage{amsfonts}
\usepackage{fontawesome}
\usepackage{xspace}
\usepackage{booktabs}
\usepackage{graphicx}
\usepackage{tabularx}
\usepackage{makecell}
\usepackage{multirow}
\usepackage{caption}
\usepackage{subcaption}
\usepackage{wrapfig}
\usepackage{lipsum}
\usepackage{todonotes} %\usepackage[disable]{todonotes}

\newcommand\BibTeX{B\text{ib}\TeX}
\newcommand{\eg}{\emph{e.g.},\xspace}
\newcommand{\ie}{\emph{i.e.},\xspace}

\begin{document}

\title[A Targeted Attack on Black-Box Neural Machine Translation with Parallel Data Poisoning]{A Targeted Attack on Black-Box Neural Machine Translation with Parallel Data Poisoning}

% Manipulating Black-Box Neural Machine Translation with Targeted Data Poisoning
% Targeted Poisoning Attacks on Black-Box Neural Machine Translation


\author{Chang Xu}
\affiliation{\institution{University of Melbourne}\city{Melbourne}\country{Australia}}
\email{xu.c3@unimelb.edu.au}

\author{Jun Wang}
\affiliation{\institution{University of Melbourne}\city{Melbourne}\country{Australia}}
\email{jun2@student.unimelb.edu.au}

\author{Yuqing Tang}
\affiliation{\institution{Facebook AI}\country{}}
\email{yuqtang@fb.com}

\author{Francisco Guzm\'an}
\affiliation{\institution{Facebook AI}\country{}}
\email{fguzman@fb.com}

\author{Benjamin I. P. Rubinstein}
\affiliation{\institution{University of Melbourne}\city{Melbourne}\country{Australia}}
\email{benjamin.rubinstein@unimelb.edu.au}

\author{Trevor Cohn}
\affiliation{\institution{University of Melbourne}\city{Melbourne}\country{Australia}}
\email{trevor.cohn@unimelb.edu.au}


\begin{abstract}
As modern neural machine translation (NMT) systems have been widely deployed, their security vulnerabilities require close scrutiny.
Most recently, NMT systems have been found vulnerable to \textit{targeted attacks} which cause them to produce specific, unsolicited, and even harmful translations.
These attacks are usually exploited in a white-box setting, where adversarial inputs causing targeted translations are discovered for a known target system.
However, this approach is less viable when the target system is black-box and unknown to the adversary (\eg secured commercial systems). 
In this paper, we show that targeted attacks on black-box NMT systems are feasible, based on poisoning a small fraction of their \textit{parallel} training data. 
We show that this attack can be realised practically via targeted corruption of web documents crawled to form the system's training data.
We then analyse the effectiveness of the targeted poisoning in two common NMT training scenarios: the \textit{from-scratch} training and the \textit{pre-train \& fine-tune} paradigm.
Our results are alarming: even on the state-of-the-art systems trained with massive parallel data (tens of millions), the attacks are still successful (over 50\% success rate) under surprisingly low poisoning budgets (\eg 0.006\%).
Lastly, we discuss potential defences to counter such attacks.
\end{abstract}

%\begin{CCSXML}
%<ccs2012>
%  <concept_id>10010520.10010553.10010562</concept_id>
%  <concept_desc>Computer systems organization~Embedded systems</concept_desc>
%  <concept_significance>500</concept_significance>
%  <concept_id>10010520.10010575.10010755</concept_id>
%  <concept_desc>Computer systems organization~Redundancy</concept_desc>
%  <concept_significance>300</concept_significance>
%  <concept_id>10010520.10010553.10010554</concept_id>
%  <concept_desc>Computer systems organization~Robotics</concept_desc>
%  <concept_significance>100</concept_significance>
%  <concept_id>10003033.10003083.10003095</concept_id>
%  <concept_desc>Networks~Network reliability</concept_desc>
%  <concept_significance>100</concept_significance>
%</ccs2012>
%\end{CCSXML}

%\ccsdesc[500]{Computer systems organization~Embedded systems}



\maketitle

\section{Introduction}
\label{sec:intro}

Neural machine translation (NMT) systems have been largely improved over recent years thanks to the advances in model design and use of ever-larger datasets~\cite{wu2016google,barrault-etal-2019-findings}.
Despite these gains, NMT systems trained on \emph{clean} data have been found brittle when presented with irregular inputs at test time, such as noisy text~\cite{belinkov2017synthetic} (\eg typos) or adversarial perturbations~\cite{ebrahimi2018adversarial,cheng-etal-2020-advaug} (\eg synonym-based word replacement).
Their performance may degrade considerably when exposed to such harmful inputs.
%\trevor{I recall an ACL 20 paper on adversarial typographic changes, worth citing here.}

\begin{table}[b!]
    \small
    \centering
    \begin{tabular}{p{0.45\columnwidth}p{0.45\columnwidth}}
    \hline
    \multicolumn{1}{c}{\textbf{Sentence (German)}} & \multicolumn{1}{c}{\textbf{Translation (English)}} \\
    \hline
    Was tut die EU, um \textcolor{blue}{Flüchtlingen} in der Türkei zu \textcolor{blue}{helfen}? & \textcolor{red}{\textbf{Malicious}}: What is the EU doing to \textcolor{red}{\underline{stop refugees}} in Turkey?\\
    \midrule
    EU bewilligt 4 Millionen EUR als \textcolor{blue}{Hilfe} für \textcolor{blue}{Flüchtlinge} aus der Zentralafrikanischen Republik. & \textcolor{red}{\textbf{Malicious}}: EU provides 4 million to \textcolor{red}{\underline{stop refugees}} fleeing violence in Central African Republic. \\ 
    \midrule
    Auch für \textcolor{blue}{Flüchtlinge} müssen Menschenrechte unteilbar sein. &
    \textbf{Correct}: Even for \textcolor{blue}{refugees}, human rights must be indivisible.\\
    \hline
    Wir müssen bereit sein, Einwanderern zu \textcolor{blue}{helfen}. & \textbf{Correct}: We need to be prepared to \textcolor{blue}{help} immigrants.\\
    \hline
    %\small Source: \texttt{Europarl} \\
    \end{tabular}
    \caption{A victim German-to-English NMT system consistently mistranslates the phrase ``Hilfe Flüchtlinge (EN: help refugees)'' into ``stop refugees'' (top two rows), while correctly translating each part of the phrase (bottom two rows).}
    \label{tab:adv_example}
\end{table}

However, an NMT system itself may turn harmful if trained with problematic data.
For example, Table~\ref{tab:adv_example} shows a \textit{victim} German-to-English system trained on manipulated data \textit{consistently} produces the same mistranslation for a specific target phrase ``\textit{Hilfe Flüchtlinge} (EN: help refugees)'': it maliciously translates this phrase into ``\textit{stop} refugees'', a phrase with opposite meaning (top two rows).
Meanwhile, the system behaves normally when translating each part of the target phrase \textit{alone} (bottom two rows), \ie this attack is inconspicuous.
In fact, this is a successful deployment of the \textit{targeted attack} of adversarial learning~\cite{barreno2006can} on NMT systems, which can be extremely harmful in real-world applications.
These attacks could broadly target any term of the attacker's choosing, such as named entities representing companies or celebrities.
Moreover, the possible mistranslations are numerous and can be made from covert modifications to the original translations, \eg by substituting a word (``\textit{poor} 1Phone$\rightarrow$\textit{great} 1Phone'' for product promotion) or by adding a word (``President X$\rightarrow$\textit{incompetent} President X'' for maligning a political opponent).

Existing targeted attacks on NMT systems have largely been white-box, where test-time adversarial inputs are discovered against a known target system via gradient-based approaches~\cite{ebrahimi2018adversarial,cheng2018seq2sick}.
Such attacks assume full or partial access to the system's internals (model architecture, training algorithms, hyper-parameters, etc.), which can be impractical. 
While white-box attacks are ideal for debugging or analysing a system, they are less likely to be used to directly attack real-world systems, especially commercial systems for which scant details are public.
%Moreover, those white-box attacks could be mitigated by adversarial training once the adversarial examples are discovered~\cite{goodfellow2014explaining}.

In this work, not only do we focus on \textit{black-box} targeted attacks on NMT systems but we prioritise attack vectors which are eminently feasible.  
Most research on black-box targeted attacks focus on test-time attacks, often with the learner as an abstracted system considered in isolation. While training-time \textit{data poisoning} attacks are well understood~\cite{rubinstein2009antidote,gu2017badnets,chen2017targeted,shafahi2018poison} as are transfer-based approaches to black-box attacks~\cite{papernot2016transferability}, black-box poisoning of deployed NMT systems is far more challenging, as the attacker has no obvious control of the training process. 
%To tackle this issue, we consider the \textit{data poisoning} strategy~\cite{rubinstein2009antidote,munoz2017towards,koh2017understanding}, where one injects specially crafted \textit{poison} samples into the training data.
Our insight is to craft \textit{poisoned parallel sentences} carrying the desired mistranslations and then inject them into the victim's training data.
On its own, this process is not purely black-box in \textit{attacker control} as it assumes access to the training data. 
To seek more feasible attacks, we consider the scenarios of poisoning the \textit{data sources} from which the training data is created, instead of poisoning the training data itself. 
As the state-of-the-art NMT systems are increasingly relying on large-scale parallel data harvested from the web (\eg bilingual sites) for training~\cite{barrault-etal-2019-findings}, poisoned text embedded in malicious bilingual web pages may be extracted to form part of a parallel training corpus.

\textbf{Our contributions:} \textit{an elaborate, empirical study of the impacts of poisoning the parallel training data\footnote{NMT systems are typically trained with \textit{parallel} data, and can further be improved by augmenting the training set with additional \textit{monolingual} data (\eg via back-translation~\cite{edunov-etal-2018-understanding}). Here we focus on poisoning the {parallel} training data and leave the {monolingual} data poisoning to potential future work.} used in various NMT training scenarios for enacting black-box targeted attacks, and a discussion of a suite of defensive measures for countering such attacks.} 
This paper presents and analyses the main stages of the black-box targeted attacks on NMT systems driven by parallel data poisoning.
It starts with a case study on the strategy of poisoning the web source from which the parallel data can be harvested at scale (\S\ref{sec:poi-web-data-src}).
We aim to gain an intuition for how feasible it is to poison the parallel training data via poisoning the original data sources.
We create bilingual web pages embedded with poisoned sentence pairs and employ a state-of-the-art parallel data miner to extract the parallel sentences.
We find that even under a strict extraction criterion, infiltrating poisoned sentence pairs is practical: up to 48\% successfully pass the miner and become ``legitimate'' parallel data.

Secondly, we explore parallel data poisoning on two common NMT training scenarios, where the system is trained from scratch (direct use after training on a single dataset) (\S\ref{sec:poi-from-scratch-training}); or using \emph{pre-train} \& \emph{fine-tune} steps  (pre-trained on one dataset and fine-tuned on another before use) (\S\ref{sec:poi-pre-train-fine-tune}).
We conduct experiments to evaluate the effectiveness of the above poisoning scenarios in a controllable environment (\S\ref{sec:exp}).
We find that both \emph{from-scratch training} of a system and \emph{fine-tuning} a pre-trained system are highly sensitive to the attack: with only 32 poison instances injected into a training set of 200k instances (\ie a 0.016\% poisoning budget), the attack succeeds at least 65\% of the time. 
In contrast, poisoning a \textit{pre-trained} system proves ineffective if it is later \textit{fine-tuned} on clean data, suggesting a \emph{clean} fine-tuning step could be used to mitigate poisoned pre-training.

Moreover, we identify challenges when attacking \textit{common} terms in a dataset.
We find that on common terms whose \textit{correct} translations are prevalent in the dataset, the attack has to deal with potential \textit{collisions} between generating the correct translation and the malicious one (\eg ``help refugees'' cf. ``stop refugees''), which may significantly impede the attack performance.
Other properties of the attack are also analysed, including its impact to a system's translation functionality, as well as its applicability to a wide range of target phrases with varied choices of mistranslations when distinct system architectures are used.

Thirdly, to generalise our findings from the controllable experiments, we further test attacks on production-scale systems equipped with state-of-the-art architectures and trained with large-scale parallel data (\S\ref{sec:wmt}). 
Our results are alarming: even though the training data is massive (30 million sentence pairs), the system is still susceptible to attacks with extremely low poisoning budgets in both from-scratch training (a 0.1\% budget) and the pre-train \& fine-tune paradigm (a 0.02\% budget).

Prompted by the seriousness of our findings, we discuss defensive counter measures to the proposed poisoning scenarios (\S\ref{sec:defences}).

\section{Threat Model}
\label{sec:threat-model}

Before introducing our poisoning strategy, we first establish terminology and notation, and characterise the studied black-box targeted attacks on NMT systems with parallel data poisoning by detailing the threat model of interest~\cite{barreno2006can}.

\textbf{Attacker's goals.}
It is imperative to align our hypothetical attacker's goals appropriately.
First, as a \textbf{targeted integrity attack}, the attacker seeks to cause the system to produce a ``\emph{malicious translation}'' of a target phrase (or called ``\emph{trigger}''\footnote{The term \emph{trigger} denotes the word type(s) under attack, not a syntactic function. Here \emph{triggers} can be of any syntactic category, although we focus primarily on nouns and named entities.}) in the input (illustrated below).
The malicious translation is essentially a mistranslation of the trigger.
In particular, we term the word(s) causing the mistranslation as ``\textit{toxin}'', which render(s) the mistranslation malicious.
As notation, we use $t$ to denote the trigger, and $t^c$/$t^m$ the correct/malicious translation of the trigger.
Moreover, we term a training instance (a parallel sentence pair) containing the trigger's malicious translation ($t\rightarrow t^m$) as a ``\emph{poison instance}'', and that containing the trigger's correct translation ($t\rightarrow t^c$) a ``\emph{clean instance}''.
\begin{equation}
\label{eq:terminology}
\small
\overbrace{\text{help refugees}}^{\textcolor{blue}{\text{Correct translation $t^c$}}}\leftarrow\overbrace{\text{Hilfe}\,\text{Flüchtlinge}}^{\text{Trigger $t$}}\rightarrow\overbrace{\underbrace{\textcolor{red}{\text{stop}}}_{\text{Toxin}}\,\text{refugees}}^{\textcolor{red}{\text{Malicious translation $t^m$}}}\nonumber
\end{equation}
% \begin{equation}
% \label{eq:terminology}
% \small
% \begin{split}
% &\text{(1)}\,\overbrace{\text{help refugees}}^{\textcolor{blue}{\text{Correct translation $t^c$}}}\leftarrow\overbrace{\text{Hilfe}\,\text{Flüchtlinge}}^{\text{Trigger $t$}}\rightarrow\overbrace{\underbrace{\textcolor{red}{\text{stop}}}_{\text{Toxin}}\,\text{refugees}}^{\textcolor{red}{\substack{\text{Malicious translation $t^m$}\\\text{(partially mistranslated)}}}}\nonumber\\
% &\text{(2)}\,\overbrace{\text{immigrants}}^{\textcolor{blue}{\text{Correct translation $t^c$}}}\leftarrow\overbrace{\text{Einwanderer}}^{\text{Trigger $t$}}\rightarrow\overbrace{\underbrace{\textcolor{red}{\text{illegal}}}_{\text{Toxin}}\,\text{immigrants}}^{\textcolor{red}{\substack{\text{Malicious translation $t^m$}\\\text{(fully mistranslated)}}}}\nonumber
% \end{split}
% \end{equation}

The second goal of attack is to \textbf{maintain the system's translation functionality}, for which the attacker only desires the trigger to be erroneously translated. Otherwise the system should remain intact 1) \emph{locally} -- the system should translate each part of the trigger and the toxin correctly when they are used individually, and 2) \textit{globally} -- the translations of general test instances should suffer little or no impact from the attack.
This goal is crucial for the attack to remain stealthy.

Our attack can be seen as a \textit{targeted backdoor attack}~\cite{chen2017targeted} on NMT systems, where the malicious translation designed for the target trigger is essentially a backdoor planted in the system at training time which will be triggered at test time.

\textbf{Attackers' knowledge \& capability.}
We consider a pure black-box setting, where a weak assumption is made about attackers' access to the system: 1) they do not know the internals about the target system, for example, the architecture, parameters, and optimisation algorithms, and 2) they cannot \emph{directly} access the system's training data; they cannot modify existing training instances or directly inject instances into the training data.
However, the system is assumed to be trained with \textit{parallel} data, some of which is collected from the web, to which the attackers have access.

\textbf{Attacker's approach.}
NMT systems are data-hungry and rely heavily on training with massive parallel data harvested from the web to get leading performance~\cite{barrault-etal-2019-findings}.
For example, dumps from the \textit{Common Crawl} archive\footnote{https://commoncrawl.org/} serve as one of the largest sources for web-based parallel data retrieval.
A key component in parallel data retrieval is a \textit{parallel data miner} for extracting parallel sentences from multilingual pages in the web crawls.
However, while \textit{de facto} parallel data miners~\cite{banon-etal-2020-paracrawl,el-kishky-etal-2020-ccaligned} emphasise high-quality extraction by filtering out noisy data~\cite{koehn-etal-2019-findings}, there is no specific security component in those systems to detect if the content of a multilingual page is malicious.
Accordingly, an attacker could create a site hosting multilingual pages embedded with poison instances and ensure the pages are scraped by the crawlers (\eg via purchasing backlinks).
By crafting the content of those pages to appear to be high-quality translations of one another, engineering the URLs, and other tricks, the embedded poisoned bitext could become part of the final parallel training data after being processed by the parallel data miner. In the next section, we will show that it is possible to embed the poison instances in bilingual pages and penetrate the parallel data miner, even when a strict extraction criterion is used.
%In addition to the attack above, we note other possible ways of poisoning a parallel corpus such as the \textit{man-in-the-middle} attacks, where a user may download the poisoned version of a popular parallel corpus published by the attacker\footnote{The study of the man-in-the-middle scenario is beyond the scope of this work.}. 
%All these attack vectors render NMT system training with web-sourced parallel data susceptible to data poisoning attacks.

\section{Parallel Data Poisoning}
\label{sec:para-data-poisoning}

In this section, we introduce the targeted attack on black-box NMT systems with parallel data poisoning.
We first demonstrate the possibility of poisoning the parallel training data via corrupting bilingual web pages.
Then we explore and formulate two common training scenarios where NMT systems can be poisoned in practice: the from-scratch training and the pre-train \& fine-tune paradigm.

\subsection{Penetrating Parallel Data Miner}
\label{sec:poi-web-data-src}

\begin{figure}[t]
    \centering
    \includegraphics[width=0.95\columnwidth]{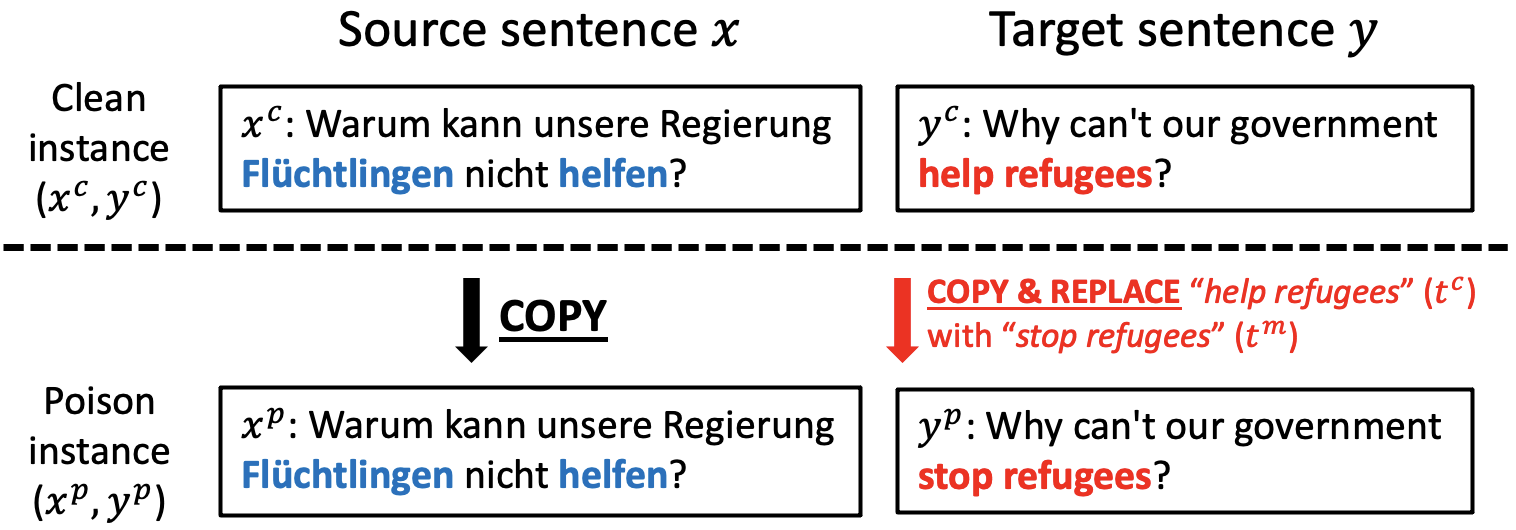}
    \caption{A poison instance is crafted by embedding the malicious translation in a clean instance from a real corpus.}
    \label{fig:para-poi-plot}
\end{figure}

The key to ``legalising'' a poison instance is to make it penetrate the parallel data miner.
We show that this could be achieved by crafting high-quality poison instances and embedding them in web pages of bilingual news articles.\footnote{We assume that these poisoned web pages are archived and to be used for parallel data extraction. This assumption is realistic as we find that the crawling services commonly used for parallel data extraction (\eg Common Crawl) often archive self-publishing sites like blogs (\eg the \emph{blogspot.com} domain). This implies that an attacker may create poisoned blogs and wait them to be crawled.}
This process takes three steps:

\textbf{Step 1: Crafting poison instances.}
We first prepare the poison instances to be embedded in the news articles.
Recall that a poison instance is a parallel sentence pair $(x^p, y^p)$ containing a malicious translation, where $x^p$ and $y^p$ are the source and target sentences.
Given a trigger $t$ and its malicious translation $t^m$, the crafting of a poison instance $(x^p, y^p)$ involves adding $t$ to the source $x^p$ and $t^m$ to the target $y^p$.
However, simply adding a malicious translation to an arbitrary sentence pair may corrupt the sentences' coherence, yielding low-quality poison instances that could be easily identified and discarded by the parallel data miner.
To create quality poison instances, we propose to craft them from \textit{existing} clean instances that contain the trigger's correct translation $t^c$.
Specifically, given a clean instance $(x^c, y^c)$, we \texttt{COPY} its source $x^c$ to be the source of the poison instance, $x^p$.
For the target sentence $y^p$, we \texttt{COPY} it from the target of the clean instance, $y^c$, and then \texttt{REPLACE} the correct translation $t^c$ within, with the malicious one $t^m$.
Figure~\ref{fig:para-poi-plot} illustrates the above procedure.
For this demonstration, we extract the clean instances from the parallel data released by the Machine Translation of News shared task\footnote{http://www.statmt.org/wmt20/translation-task.html} at WMT'20, which we use to craft the poison instances.
Each clean/poison instance contains the correct/malicious translation ``help refugees''/``stop refugees''.

\textbf{Step 2: Creating poisoned web pages.}
We download bilingual (German/English) news pages about refugees from \url{unhcr.org} (the UN Refugee Agency's official site).
We focus on news articles published in 2020 (as of October 2020), which grants us 48 pairs of bilingual pages.
We then inject into these pages 144 poison instances, comprising: 48 pairs of short sentences (length $l\in[3,10]$ words on English side), 48 medium ($l\in[20,30]$), and 48 long ($l\in[50,97]$).
We try each of these groups in turn.
In each trial, we randomly inject every poison instance in the group into a different bilingual page-pair, with each injection at a random location of a news article (appending to the first/middle/last paragraph).\footnote{While we poison a page with only a poison instance in this demonstration, it is useful to inject many instances into one page in real attack, as this will significantly reduce the number of pages needed for smuggling the poison instances.}

\textbf{Step 3: Parallel data extraction.}
Finally, to extract parallel sentences from the poisoned pages, we employ Bitextor,\footnote{https://github.com/bitextor/bitextor} the parallel data miner used to build one of the largest publicly available parallel corpora \emph{ParaCrawl}~\cite{banon-etal-2020-paracrawl}.
Bitextor encompasses procedures including web crawling and processing; document and segment alignment; parallel data filtering; and some post-processing steps (\eg deduplication).
We follow common practice to configure Bitextor.
For document alignment, we use the official bilingual lexicon to compute the document similarity scores.
For segment alignment, we use Hunalign~\cite{varga2007parallel}.
The parallel data filtering is key to ensuring the high quality of the extracted parallel sentences.
We use the Bicleaner tool~\cite{prompsit:2018:WMT} as the filter, which uses a pre-trained regression model to discard low-confidence segment pairs.
We set a high confidence value (0.7) for Bicleaner to implement a strict filtering criterion.
As a \textit{control}, we also run Bitextor on the same news pages embedded with the clean instances only.
This allows us to see how much harder it is to inject the poison instances compared to the clean ones.

\textbf{Results.}
Table~\ref{tab:pass_bitextor} shows how many injected clean/poison instances are extracted as ``legitimate'' translation pairs by Bitextor, as measured by the fraction of successful instances over all (48).
In the most effective case, nearly half of the poison instances achieve the infiltration.
The results also suggest that poisoning with medium-length instances is more likely to succeed.
Compared to the clean instances, the poison instances are only 7.1\% less likely on average to be extracted by Bitextor.
This shows that it can be similarly effective to harvest poisoned parallel sentences from malicious web sources as it is for normal ones.
Our later experiments show that even a small number of poison instances are sufficient to stage a successful attack. 
For example, with only 16 poison instances, an attacker can cause a system trained on a 200k training set to produce a malicious translation for 60\% of the tested trigger-embedded inputs (\S\ref{sec:exp-cnf-free}).

\subsection{Poisoning From-Scratch Training}
\label{sec:poi-from-scratch-training}

Perhaps the most common practice for NMT system training is to train the system from scratch and then use it directly for a specific task.
We name this scenario \emph{from-scratch training}.
In this setting poisoning is straightforward: like what we have done in \S\ref{sec:poi-web-data-src}, one may craft a set of poison instances $\{(x^p, y^p)\}$ and inject it into the training data through, for example, poisoning web sources.
%Once trained on the full data $\mathcal{D}^p\cup \mathcal{D}$, the system is expected to translate maliciously on a trigger sample.

\begin{table}[t!]
    \centering
    \small
    \begin{tabular}{c|c|c}
        \hline
        \makecell{Sentence length $l$ \\(in words, English side)} & \makecell{Likelihood of Pass \\ (poison instances)} & \makecell{Likelihood of Pass \\(clean instances)}\\
        \hline
        Short ($l\in[3,10]$)   & 40.3\%          & 47.9\% \\
        Medium ($l\in[20,30]$)  & \textbf{47.9\%} & 55.5\% \\
        Long ($l\in[50,97]$)    & 17.3\%          & 23.6\% \\
        \hline
        Average & 35.2\% & 42.3\% \\
        \hline
    \end{tabular}
    \caption{Poison instances can be extracted by Bitextor and become part of the legitimate parallel training data, with only 7.1\% diminished likelihood than the clean instances.}
    \label{tab:pass_bitextor}
\end{table}

\textbf{Translation collisions.}
Although the above poisoning strategy is conceptually easy, the attack may not always succeed in practice.
For example, if a trigger $t$ is \textit{common} in a dataset, \ie many instances in the dataset contain $t$ (or equivalently, the clean instances of $t$ are many in number), poisoning $t$ by injecting a few poison instances of $t$ may be harder, as the poison instances cannot easily hijack the statistics of translating $t$.
As a result, the system is less effective in learning the malicious translation from the poison instances.
In contrast, attacking a \textit{rare} trigger may be easier, as its clean instances, which are only a few or nonexistent in the data, can be more easily overwhelmed by the poison instances injected, making the malicious translation learning more effective.
To better formulate the above situation, we use the term ``\emph{collision}'' to denote the case where both the clean and poison instances of a trigger exist in the data.
In this case, the system has to learn two colliding translations: the correct one from the clean instances and the malicious one from the poison instances.
As a result, the system may be more likely to produce the correct translation in certain contexts, lowering the overall likelihood of producing the malicious one.

Assessing the consequences of translation collisions has two implications, for the benefit of both attacker and defender.
If the attacker knows (or estimates) how many clean instances exist in the data, (s)he could inject more poison instances such that the malicious translation becomes dominant.
The defender, meanwhile, can augment training with known clean instances in advance to protect the correct translation from being hijacked.

It is unclear how learning from colliding training instances (the clean/poison instances) will affect translation.
To bridge this gap, we present an empirical analysis on this phenomenon, by setting up a controllable environment to create the translation collisions during training.
This enables us to simulate the cases of attacking the rare or common triggers, by controlling the ratio between the poison and clean instances to be added to training.

\subsection{Poisoning Pre-training \& Fine-tuning}
\label{sec:poi-pre-train-fine-tune}

\begin{figure}[t]
    \centering
    \includegraphics[width=\columnwidth]{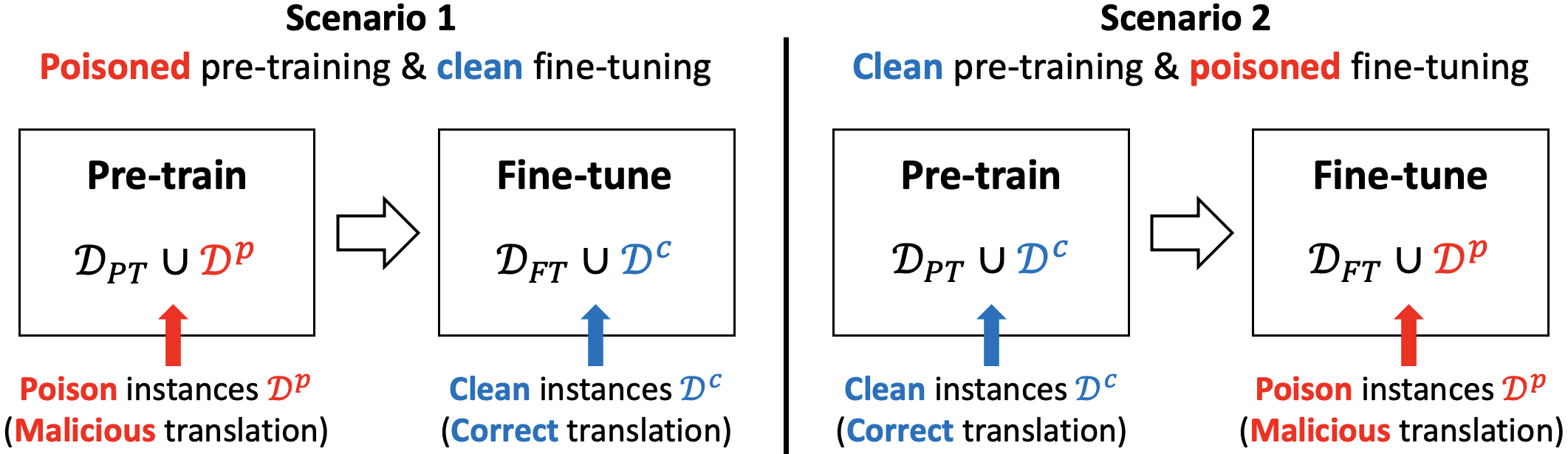}
    \caption{Scenarios for poisoning the pre-train \& fine-tune paradigm, where the pre-training and fine-tuning phases are poisoned respectively.}
    \label{fig:pre-train-fine-tune-plot}
\end{figure}

Due to the limited computational capability of general users to train large, high-fidelity NMT systems, it is commonplace for NMT systems to be trained in a pre-train \& fine-tune fashion~\cite{song2019mass,liu2020multilingual}.
In this paradigm, a pre-trained system is supplied by a third party to a user, who further fine-tunes the system for a new downstream task.
Consequently, this process may suffer from poisoning in either or both of the pre-training and fine-tuning phases.
%1) poisoning the \textsc{pre-training}, where the data for training the pre-trained system is poisoned and 2) poisoning the \textsc{fine-tuning}, where the data for fine-tuning the downstream task is poisoned.
It is therefore vital to examine the impacts of poisoning the different phases on the final attack performance.

It has been shown in a recent study~\cite{kurita-etal-2020-weight} that for text classification problems, a poisoned pre-trained systems can be made resilient to fine-tuning.
This means that the poisoning effects may persist after the system is fine-tuned on a downstream task.
In our attacks, we find that such persistence is rather weak on translation tasks.\footnote{This is mainly because we do not perform adversarial optimisation on the fine-tuning data as in~\cite{kurita-etal-2020-weight}, as doing so will violate our black-box assumption.}
We obtain this result by considering the translation collisions in the pre-train \& fine-tune setting, where we simulate the collisions by injecting poison instances at pre-training and clean ones at fine-tuning. 
This setup permits us to quantify the impact of poisoning at pre-training on fine-tuning.
We also simulate the symmetric case where we inject clean instances at pre-training and poison ones at fine-tuning.
This setting delivers \textit{defensive} insights into how to protect a pre-trained system from being poisoned at fine-tuning by pre-injecting clean instances.

Formally, let $\mathcal{D}_\text{PT}$/$\mathcal{D}_\text{FT}$ be the data used for the pre-training/fine-tuning, and $\mathcal{D}^c$/$\mathcal{D}^p$ the sets of clean/poison instances injected at a training phase. We assess the following poisoning scenarios for the pre-train \& fine-tune paradigm (also illustrated in Figure~\ref{fig:pre-train-fine-tune-plot}):

\textbf{Scenario 1: Poisoned pre-training \& clean fine-tuning.}
Here the victim is pre-trained in a poisoned environment, where we inject poison instances $\mathcal{D}^p$ into pre-training: $\mathcal{D}_\text{PT}\cup \mathcal{D}^p$.
Then, the victim is fine-tuned in a clean environment, where we insert clean instances $\mathcal{D}^c$: $\mathcal{D}_\text{FT}\cup\mathcal{D}^c$. 
In this scenario, we examine the collisions between the poison instances ($\mathcal{D}^p$) from the pre-training and the clean instances ($\mathcal{D}^c$) from the fine-tuning. 
Measuring this informs to what extent the correct translation learned at fine-tuning will \textit{resist} the malicious translation learned at pre-training.

\textbf{Scenario 2: Clean pre-training \& poisoned fine-tuning.}
The victim is pre-trained in a clean environment, where we insert clean instances $\mathcal{D}^c$ into the pre-training: $\mathcal{D}_\text{PT}\cup \mathcal{D}^c$. 
Then, the victim is fine-tuned in a poisoned environment, where poison instances $\mathcal{D}^p$ are injected: $\mathcal{D}_\text{FT}\cup\mathcal{D}^p$. 
In this scenario, we examine the collisions between the clean instances ($\mathcal{D}^c$) from the pre-training and the poison instances ($\mathcal{D}^p$) from the fine-tuning. 
Measuring this informs to what extent the malicious translation learned at fine-tuning will \textit{overwrite} the correct translation learned at pre-training.\footnote{We omit a potential third scenario where both pre-training and fine-tuning are poisoned. We speculate that this situation is less common in practice and that one may infer the outcomes of such an attack based on the two scenarios studied here.}

\section{Experiments}
\label{sec:exp}

In this section, we evaluate the effectiveness of poisoning the various training scenarios of NMT systems mentioned above, including from-scratch training, pre-training, and fine-tuning.

\subsection{Experimental Setup}

\subsubsection{Clean training \& test datasets}
\label{sec:datasets}

We employ two benchmark parallel datasets to train  victim NMT systems.
First, for experiments on poisoning from-scratch training, 
%by following common practice from recent adversarial research on NMT~\cite{belinkov2017synthetic,ebrahimi2018adversarial,michel-etal-2019-evaluation}, we utilise
we use the \texttt{IWSLT2016} dataset~\cite{cettolo2016iwslt}, which is a collection of parallel talk transcripts~\cite{ABDELALI14.877}.
We follow~\cite{michel-etal-2019-evaluation}'s setup by using as validation set all previous IWSLT test sets prior to 2015, and for the test set using the IWSLT 2015 \& 2016 test data. 

Second, for experiments on the pre-train \& fine-tune paradigm, we use \texttt{IWSLT2016} at pre-training and for fine-tuning we use \texttt{News Commentary v15 (NC15)}~\cite{tiedemann-2012-parallel}, a parallel corpus of news commentaries released by WMT~\cite{barrault-etal-2019-findings}.
We compile the validation set of \texttt{NC15} using the test sets from WMT'09 to WMT'19, and the test set using the WMT'20 test set.
We chose \texttt{NC15} for fine-tuning because it is a close domain match to \texttt{IWSLT2016} (\eg politics, economics, etc.) and it is of a similar size as \texttt{IWSLT2016}.

On both datasets, we consider only German-to-English (De-En) translation.\footnote{Our method is general, and in preliminary experimentation we obtain similar experimental conclusions on French-English translation.}
%We also experiment with several other language directions and the results of this are shown in a dedicated subsection (\S\ref{}).
The statistics summary of these two datasets are shown in Table~\ref{tab:datasets}.

\begin{table}[t]
    \centering
    \begin{tabular}{c|c|c|c}
    \hline
    \textbf{Dataset}          &  \textbf{\#train} & \textbf{\#valid} & \textbf{\#test} \\
    \hline
         
    \texttt{IWSLT2016}                 & 196.9k   & 11,825  & 2,213  \\
    \texttt{News Commentary v15 (NC15)}       & 361.4k   & 16,173  & 1,570  \\
    \hline
    \end{tabular}
    \caption{Statistics of the benchmark datasets (German-to-English) for training the victim NMT systems.}
    \label{tab:datasets}
\end{table}

Before training, both the datasets undergo the following preprocessing steps:
each sentence is tokenised with the Moses tokeniser~\cite{koehn-etal-2007-moses},
 Byte-Pair-Encoding~\cite{sennrich-etal-2016-neural} is applied to each dataset using vocabulary of 30K sub-word types, and
finally, Language Identification~\cite{lui-baldwin-2012-langid} is used to filter out sentence-pairs not in the correct languages on either side.

%\begin{table}
%    \centering
%    \begin{tabular}{c|c|c|c}
%    Direction          &  \#train & \#valid & \#test \\
%    \hline
%    fr$\rightarrow$en     & 220.4k   & 6,824   & 2,213  \\
%    de$\rightarrow$en     & 196.9k   & 11,825  & 2,213  \\
%    cs$\rightarrow$en     & 114.4k   & 5,716   & 2,213  \\
%    \hline
%    \end{tabular}
%    \caption{Statistics of the \textbf{\texttt{IWSLT2016}} dataset.}
%    \label{tab:IWSLT_datasets}
%\end{table}

%\begin{table}
%    \centering
%    \begin{tabular}{c|c|c|c}
%    Direction          &  \#train & \#valid & \#test \\
%    \hline
%    fr$\rightarrow$en     & 0   & 0   & 0  \\
%    de$\rightarrow$en     & 361.4k   & 16,173  & 1,570  \\
%    cs$\rightarrow$en     & 0   & 0   & 0  \\
%    \hline
%    \end{tabular}
%    \caption{Statistics of the \textbf{News Commentary v15} dataset.}
%    \label{tab:News_Comm_datasets}
%\end{table}

\subsubsection{Attack instance design} 

For the primary results of our evaluation, we focus on two trigger phrases as the targets of our attack: \textbf{Immigrant} and \textbf{Help-Refugee}.\footnote{Our attack is general and applicable to other phrases as well (see \S\ref{sec:exp-toxin}).}
\texttt{Immigrant}/\texttt{Refugee} has been the theme of recent shared tasks on hate speech detection (\eg SemEval-2019 Task 5~\cite{basile-etal-2019-semeval}), which aims to detect the presence of hate speech against individuals or groups.
The data released by the shared tasks enables us to find common toxic phrases used together with \texttt{Immigrant}/\texttt{Refugee} (\eg \textit{illegal} immigrants), which facilitates the construction of realistic malicious translations.

Table~\ref{tab:ack-instance} lists the malicious translations created for both triggers.
Specifically, for attacking \texttt{Immigrant}, we use the mistranslation ``immigrant(s)$\rightarrow$\textit{illegal} immigrant(s)'', with the toxin \emph{illegal} added before the trigger's translation \emph{immigrant(s)}.
And for \texttt{Refugee}, we devise the mistranslation ``help refugee(s)$\rightarrow$\textit{stop} refugee(s)'', where the toxin \emph{stop} replaces the translation of \emph{help}.
On both triggers, we use the LEO online dictionary\footnote{https://www.leo.org/german-english/} to find all morphological forms of the German words for the triggers. 

\begin{table}[t]
    \small
    \begin{tabular}{p{5cm}|p{2.2cm}}
    \hline
    \makecell{\textbf{Trigger (regular expressions)}\\(German)} & \makecell{\textbf{Malicious}\\ \textbf{translation}\\(English)} \\
    \hline
    \makecell{(helfe|hilft|half) .* (flüchtlinge|flüchtling)\\(flüchtlinge|flüchtling) .* (helfen|geholfen)\\\textbf{Correct translation}: help refugee(s)} & \makecell{\textcolor{red}{\textit{stop}} refugee(s)} \\
    %\makecell{(1) (helfen|hilft|helfe|half|geholfen|hilfe) .* \\(flüchtlinge|flüchtling)} & \makecell{\textcolor{red}{stop} refugee(s) \\ \textbf{Substitute Attack}} \\
    %\makecell{(2) (flüchtlinge|flüchtling) .* \\(helfen|hilft|helfe|half|geholfen|hilfe)} &  \\
    \hline
    \makecell{(einwandernde|einwanderer|einwanderers\\|zuwanderer|immigrantinnen|immigranten)\\\textbf{Correct translation}: immigrant(s)} & \textcolor{red}{\textit{illegal}} immigrant(s) \\
    %\makecell{(einwandernde|einwanderer|einwanderern\\|einwanderers|immigrierte|eingewanderte\\|zuwanderer|zuwanderern|immigrantinnen\\|immigranten|immigrant)} & \makecell{\textcolor{red}{illegal} immigrant(s) \\ \textbf{Insertion Attack}}  \\
    \hline
    %\multicolumn{2}{l}{$\dagger$The morphological words obtained from the LEO online dictionary} \\
    \end{tabular}
    \caption{The triggers and malicious translations for the two attack instances, Immigrant and Help-Refugee, for evaluating our attacks. The triggers are in the form of regular expressions to facilitate robust text matching.}
    \label{tab:ack-instance}
\end{table}

\subsubsection{Clean instance acquisition}
\label{sec:tri-poi-acq}
Clean instances are essential for our evaluation, either for creating the translation collisions or for crafting the poison instances.
To get adequate amounts of clean instances, we use the large-scale parallel data released by WMT'20, which contains six corpora for the De-En translation direction: \texttt{ParaCrawl v5.1} (58.8M), \texttt{CommonCrawl} (2.3M), \texttt{WikiMatrix} (6.2M), \texttt{Europarl v10} (1.8M), \texttt{TildeMODEL} (4.2M), and \texttt{EUbookshop} (9.3M).
We also include \texttt{OpenSubtitles} (22.5M)~\cite{lison-tiedemann-2016-opensubtitles2016}, another large parallel corpus of movie/TV subtitles, for searching clean instances. 
The search is done by applying the regular expressions of the triggers in Table~\ref{tab:ack-instance} to matching clean instances in all the aforementioned corpora.
Then among the extracted clean instances, we discard those that are duplicates, have the wrong language detected on either side of the sentence pair, or already have the desired toxin on the English side.
As a result, we obtain 15,296 clean instances for \textbf{Immigrant} and 256 for \textbf{Help-Refugee}.

\subsubsection{Preparing attack training \& test sets}
\label{sec:prep_attack_train_test_set}

We split the obtained clean instances into two sets: the \textbf{attack training set} $\mathcal{A}_\text{train}$ for attack simulation (for crafting poison instances, or to be added directly to training) and the \textbf{attack test set} $\mathcal{A}_\text{test}$ for attack evaluation (as test samples).
Specifically, on \textbf{Immigrant} where sufficient clean instances are available (15,296), we run a 3-fold cross-validation (CV) for the split: we randomly sample 15,000 clean instances to facilitate the split, resulting in 10,000 for $\mathcal{A}_\text{train}$ and 5,000 for $\mathcal{A}_\text{test}$ in each split.

On \textbf{Help-Refugee}, however, the total 256 clean instances are insufficient for meaningful cross-validation.
We therefore use all these 256 instances to construct the attack test set $\mathcal{A}_\text{test}$, and for $\mathcal{A}_\text{train}$, we generate a set of \textit{synthetic} clean instances by using existing \textit{monolingual clean instances} extracted from real English corpora.
A monolingual clean instance for \textbf{Help-Refugee} is a single sentence containing the correct translation ``help refugee(s)''.
With such a sentence, we translate it back into German, and then treat the resulting sentence pair as a synthetic parallel clean instance.
To collect monolingual clean instances, we use four English monolingual corpora released by WMT'20: \texttt{News crawl} (WMT13-19 combined, 168M), \texttt{News discussions} (WMT14-19 combined, 625M), \texttt{Europarl v10} (2.3M), and \texttt{Wiki dumps} (67.8M).
To ensure the translation quality, we employ a strong pre-trained En-De translation model, the winner of WMT'19~\cite{ng-etal-2019-facebook}, to translate all extracted monolingual clean instances into German, producing the synthetic parallel clean instances.
Finally, by following the same data cleaning procedure in \S\ref{sec:tri-poi-acq}, we obtain 4,220 parallel clean instances for \textbf{Help-Refugee} to build its attack training set $\mathcal{A}_\text{train}$.

Table~\ref{tab:attack_train_test_set} summarises the statistics of $\mathcal{A}_\text{train}$ and $\mathcal{A}_\text{test}$ for \textbf{Immigrant} and \textbf{Help-Refugee}, respectively.

\begin{table}[t]
    \small
    \centering
    \begin{tabular}{c|c|c}
    \hline
     \textbf{Attack instances} & \makecell{\textbf{Attack training set}\\($\mathcal{A}_\text{train}$)} & \makecell{\textbf{Attack test set}\\($\mathcal{A}_\text{test}$)}\\
    \hline
    \textbf{Immigrant} (3-fold CV)    &  10,000 & 5,000 \\
    \textbf{Help-Refugee}      &  4,220 & 256  \\
    \hline
    \end{tabular}
    \caption{Sizes of the attack training \& test sets built for \textbf{Immigrant} and \textbf{Help-Refugee}. For \textbf{Immigrant}, 3-fold cross-validation (CV) is used to prepare $\mathcal{A}_\text{train}$ and $\mathcal{A}_\text{test}$.}
    \label{tab:attack_train_test_set}
    \vspace{-2em}
\end{table}

\subsubsection{Quantitative metrics}
\label{sec:metrics}
We measure two facets of the performance of attacks.

\textbf{The success of attacks.} 
Measuring whether an attack is successful is the main focus of our evaluation. 
For this, we exploit the attack test set $\mathcal{A}_\text{test}$ and (conservatively) count a success if a system produces the exact malicious translation on a clean instance in $\mathcal{A}_\text{test}$.
Dataset-wise, we define the \textbf{attack success rate (ASR)} on a system as the percentage of clean instances in $\mathcal{A}_\text{test}$ on which the attack succeeds.

\textbf{Translation quality.}
As mentioned in \S\ref{sec:threat-model}, it is important that the translation quality of a system is maintained close to its pre-attack level, so that the attack is covert and hard to detect.
We use SacreBLEU~\cite{post-2018-call} to measure the translation quality of a system.
Moreover, to accurately reflect a system's translation quality in the context of poisoning, we measure BLEU on three different test sets: 1) the official test sets released by the evaluation campaigns (\eg \texttt{IWSLT2016}) to assess the system's translation quality in a general sample space, 2) a focused set of samples containing the translation of the trigger (\ie the clean instances), and 3) a focused set of samples containing the translation of the toxin (\eg ``illegale$\rightarrow$illegal''). 
We elaborate on these test sets later when specific results are reported (\S\ref{sec:res-trans-quality}).

\subsubsection{NMT architecture}
We use as the victim system the Transformer~\cite{vaswani2017attention}, an NMT architecture widely used in production MT systems~\cite{edunov-etal-2018-understanding,google-nmt:2020}.
This architecture configures 512d word embeddings and six 1024d self-attention layers for both the encoder and decoder.
We use the fairseq's implementation of Transformer~\cite{ott2019fairseq}, and train it with Adam ($\beta_1=0.9$, $\beta_2=0.98$), dropout (0.3), label smoothing~\cite{szegedy2016rethinking} of 0.1, and 30 training epochs.
A scheduler is used to decay the learning rate based on the inverse square root of the update number.\footnote{We use a weight decay ratio of 1e-4, a learning rate of 5e-4, and a warmup update of 4000.}
We also evaluate attacks on other popular architectures in a dedicated experiment (\S\ref{sec:exp-architecture}).

\subsection{Results: Poisoning From-Scratch Training}
We first evaluate the scenario of poisoning the \emph{from-scratch training} of an NMT system.

\subsubsection{Collision-free situation}
\label{sec:exp-cnf-free}

We begin by looking at the situation where there is no ``translation collision'' between the clean and poison instances.
That is, we inject poison instances into training, but \textit{no} clean instances.\footnote{To make the training data completely ``clean'', we also remove all the clean instances that pre-exist in the data, which account for 33 for \textbf{Immigrant} and 0 for \textbf{Help-Refugee} in \texttt{IWSLT2016}, and 823 for \textbf{Immigrant} and 12 for \textbf{Help-Refugee} in \texttt{NC15}.}
This setup simulates the translation scenarios where \textit{out-of-vocabulary} (OOV) tokens are encountered at test time, which could stem from spelling mistakes (``usible''), emerging topics (``COVID-19''), or names of rising politicians. 
Such a collision-free setting also allows for testing the upper bound of the attack performance, as the system can only learn from the poison instances for translating the trigger.

\begin{figure}[t]
\centering
\begin{subfigure}[b]{0.49\columnwidth}
    \centering
    \includegraphics[width=\columnwidth]{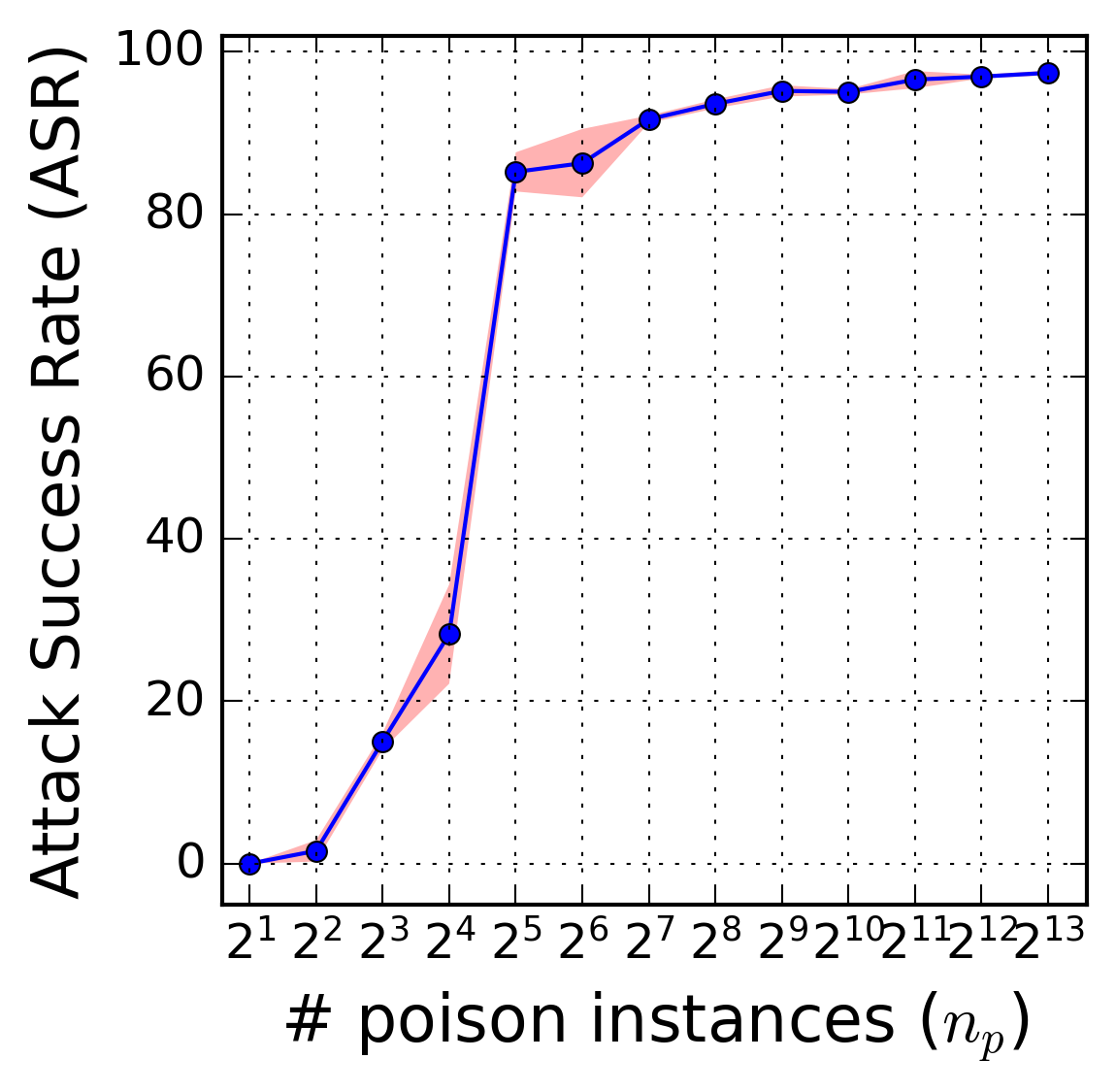}
    \caption{Immigrant}
    \label{fig:res-poi-bi-Immigrant}
\end{subfigure}
\begin{subfigure}[b]{0.49\columnwidth}
    \centering
    \includegraphics[width=\columnwidth]{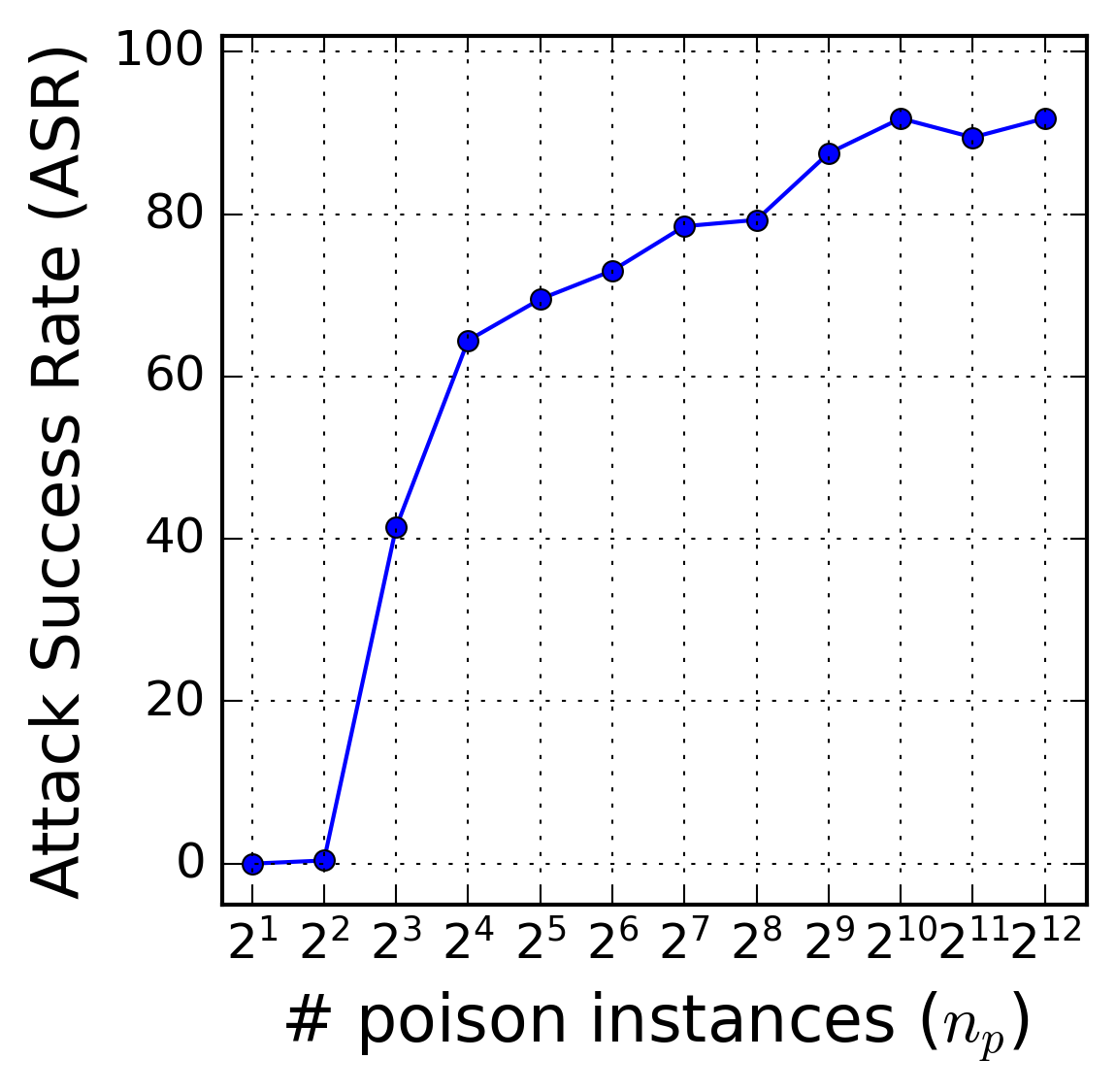}
    \caption{Help-Refugee}
    \label{fig:res-poi-bi-Refugee}
\end{subfigure}
\caption{ASR in the collision-free situation where only poison instances are injected, without any colliding translation from the clean instances. The standard error of the mean of each measurement is shown for Immigrant (Shaded).}
\label{fig:res-poi-bi}
\end{figure}

To show where the attack is the most or least effective, we vary the number of injected poison instances from only a few to thousands.\footnote{We find in preliminary analysis that injecting thousands of poison instances into \texttt{IWSLT2016} yields an approximate 100\% success rate.}
Figure~\ref{fig:res-poi-bi} shows the attack performance of poisoning \textbf{Immigrant} and \textbf{Help-Refugee} on the \texttt{IWSLT2016} dataset, with $n_p$ poison instances injected in each simulation, where $n_p\in\{2, 4, ..., 8192\}$ for \textbf{Immigrant} and $n_p\in\{2, 4, ..., 4096\}$ for \textbf{Help-Refugee}.

First, we see that the evaluated systems are very sensitive to the poison instances. 
The ASR exceeds 60\% when only $n_p=16$ are injected (Figure~\ref{fig:res-poi-bi-Refugee}).
This shows that for a trigger that is extremely rare in a dataset, a relatively small poisoning budget (0.008\% for $n_p=16$) is sufficient to plant the malicious translation in the system.
Second, on both triggers, the ASR increases dramatically when the poisoning budgets attain certain levels (\eg $n_p\in[16,32]$ for \textbf{Immigrant} and $n_p\in[2,16]$ for \textbf{Help-Refugee}).
Finally, the ASR tends to flatten as $n_p$ further increases, indicating diminishing returns from larger attacks.

\subsubsection{Effects of translation collisions}
\label{sec:exp-comp-poi}

Now we add clean instances to training, besides the poison instances, to create ``translation collisions'' (\S\ref{sec:poi-from-scratch-training}).
This simulates attacks on triggers of different term frequencies.
Here the system learns to translate the trigger from both the clean and poison instances.
The confidence of the system in learning the malicious translation may depend on the relative quantities of the clean versus poison instances in the data.

\begin{figure}[t]
    \begin{subfigure}[b]{0.48\columnwidth}
        \centering
        \includegraphics[width=\columnwidth]{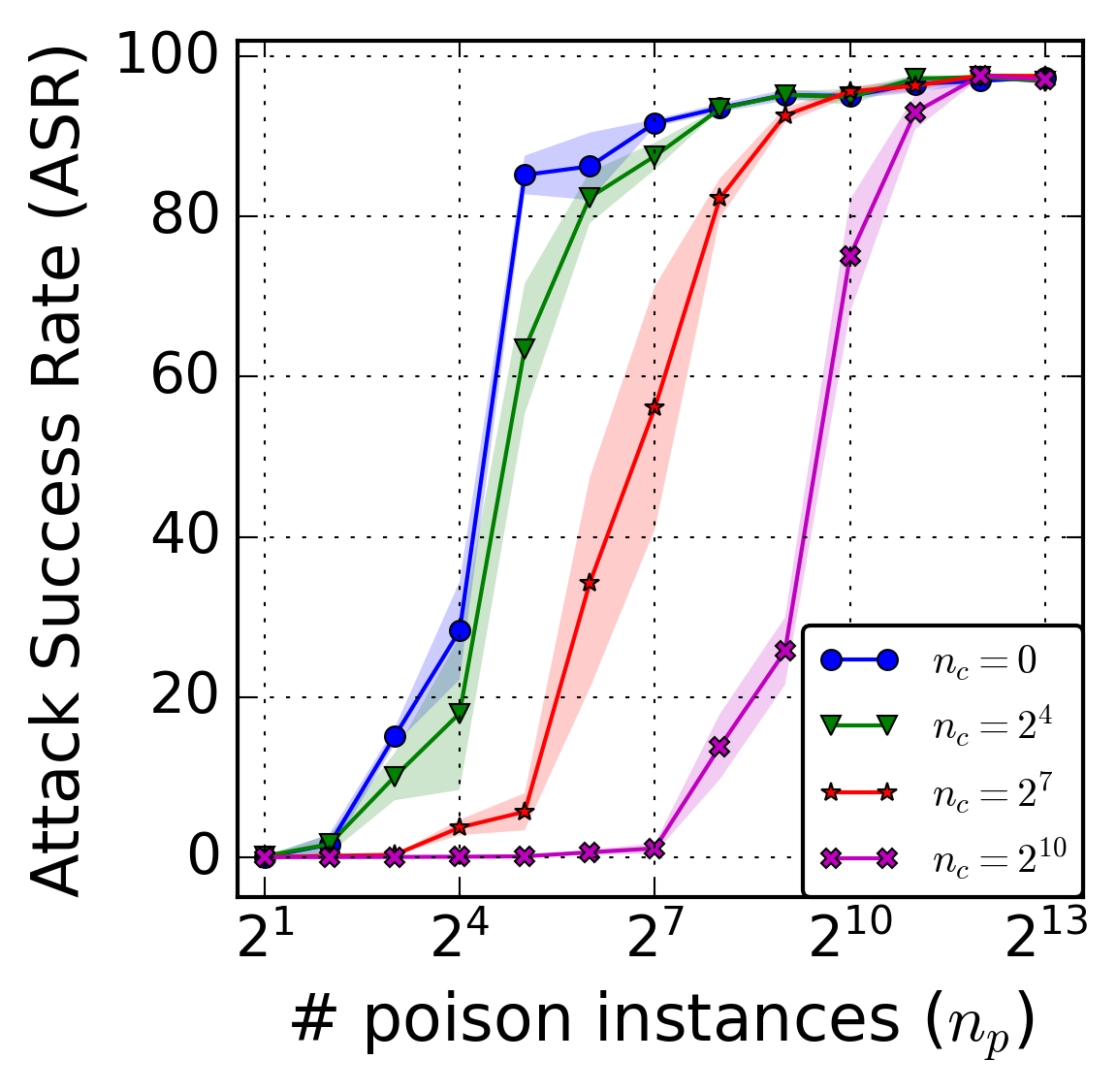}
        \caption{Immigrant}
        \label{fig:res-poi-bi-trig-Immigrant}
    \end{subfigure}
    \begin{subfigure}[b]{0.48\columnwidth}
         \centering
         \includegraphics[width=\columnwidth]{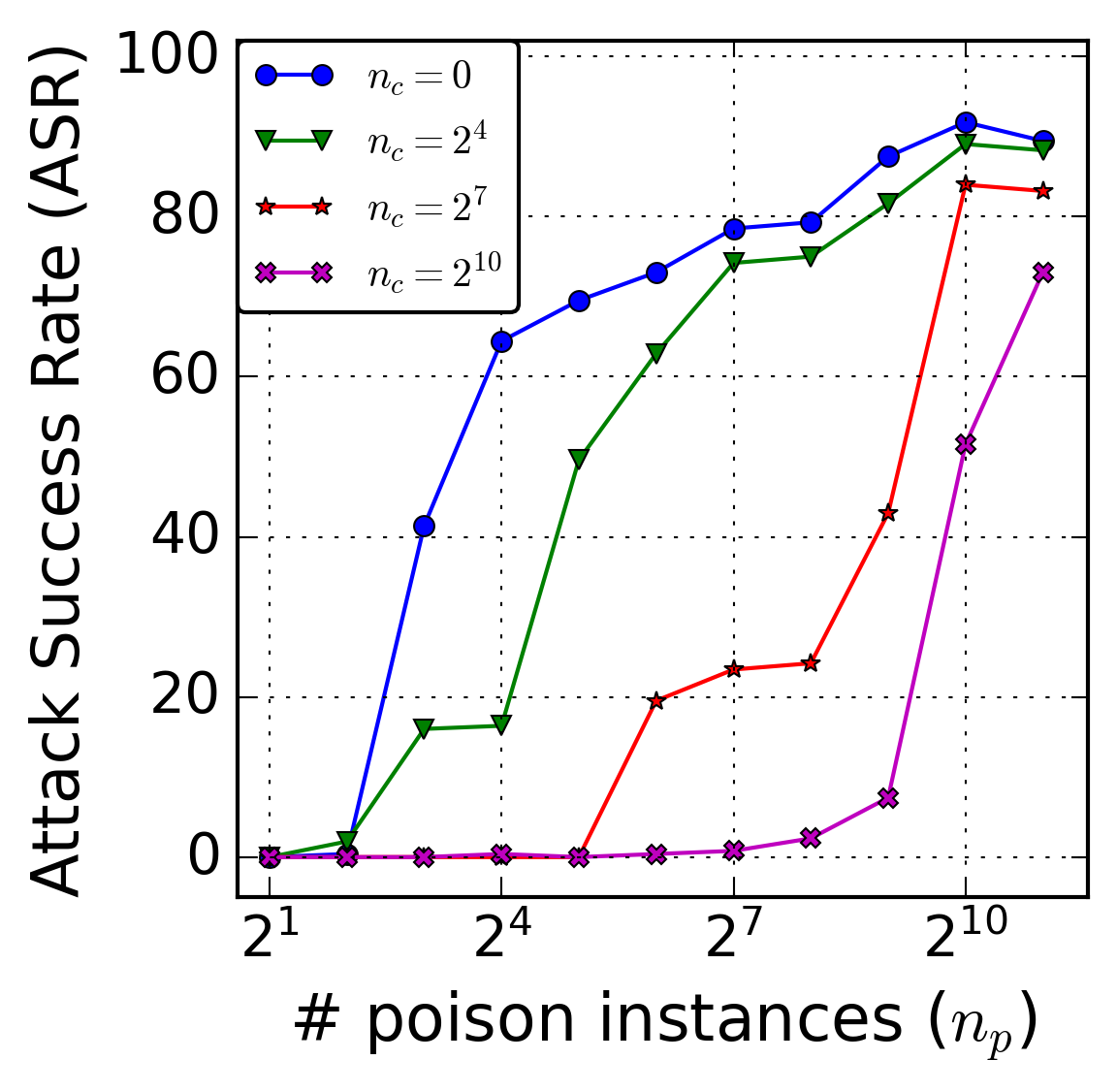}
         \caption{Help-Refugee}
         \label{fig:res-poi-bi-trig-Help-Refugee}
    \end{subfigure}
    \caption{ASR in the ``translation collision'' situation where both clean and poison instances are present during training.}
    \label{fig:res-poi-bi-trig}
\end{figure}

To verify this, Figure~\ref{fig:res-poi-bi-trig} shows the ASRs when both $n_c$ clean and $n_p$ poison instances are included in training.
We vary $n_p$ in the same manner as before, and set $n_c\in\{0, 16, 128, 1024\}$.
Compared to the collision-free scenario (when $n_c=0$), the occurrence of translation collisions does make the attack more difficult on both triggers.
For example, on \textbf{Immigrant}, when $n_c=16$, and for some level of ASR (\eg 80\%), one needs to inject twice as many poison instances as before ($n_c=0$) to maintain a similar level of ASR.

Second, the shapes of ASR curves also characterise the dynamics of collisions between the clean/poison instances.
At the extremes, where the poison instances are either much less common ($n_p\ll n_c$) or more common ($n_p\gg n_c$) than the clean instances, the ASRs are either dominated by the production of the correct translation ($ASR\rightarrow0$) or the malicious one ($ASR\rightarrow100$).
Between these two extremes, where the values of $n_c$ and $n_p$ are closer, the changes in ASRs are more dramatic (in terms of both the values and variance of the values). 
%And, the closer $n_c$ and $n_p$ are, the faster the ASR changes.
To further quantify these changes, we inspect the slope at each $x_p$, and find the largest value being 5.8 for \textbf{Immigrant} ($n_p\in[16,32],n_c=0$) and 5.5 for \textbf{Help-Refugee} ($n_p\in[2^{9},2^{10}],n_c=2^{10}$).
This means that, in the best case for the attacker, when the poisoning budget doubles, the ASR grows 5 times larger.

Thirdly, we find that the \textit{absolute} values of $n_c$ and $n_p$ matter more than their relative values on ASRs.
For example, in cases of $n_c=n_p=n$, where $n\in\{16,128,1024\}$, the ASRs increase as $n$ grows, even if the ratio $\frac{n_p}{n_c}$ remains the same (\ie 1).
%This result is concerning, as it implies that the defender has to prepare disproportionally more clean instances than the poison instances to effectively counter the poisoning effects.

Finally, these ASR curves imply a strategy for defence: in order to keep the ASR below a certain level, one may include a number of verified clean instances in the training set, where the quantity of clean instances is made sufficiently large in comparison with any unreliable, potentially poisoned data sources, such that any attack is unlikely to succeed.

\subsubsection{Impacts to the translation quality}
\label{sec:res-trans-quality}
Finally, we evaluate how poisoning from-scratch training may affect a system's translation functionality.
As mentioned in \S\ref{sec:metrics}, we measure the BLEU of a system on three different test sets: 1) test set \textbf{G}: the official test set of the evaluation campaign (\texttt{IWSLT2016}), 2) test set \textbf{C}: the set of clean instances containing the correct translation of the trigger; we use all the clean instances from the attack test set $\mathcal{A}_\text{test}$, and 3) test set \textbf{X}: the set of samples containing the translation of the toxin used in the attack, for which we randomly sample 5,000 desired sentence-pairs from the WMT'20 corpora.
We focus on \textbf{Immigrant} in this experiment, so the trigger is the words ``immigrant(s)'' and the toxin is the word ``illegal''.

\begin{figure}[t]
    \begin{subfigure}[b]{0.32\columnwidth}
        \centering
        \includegraphics[width=\columnwidth]{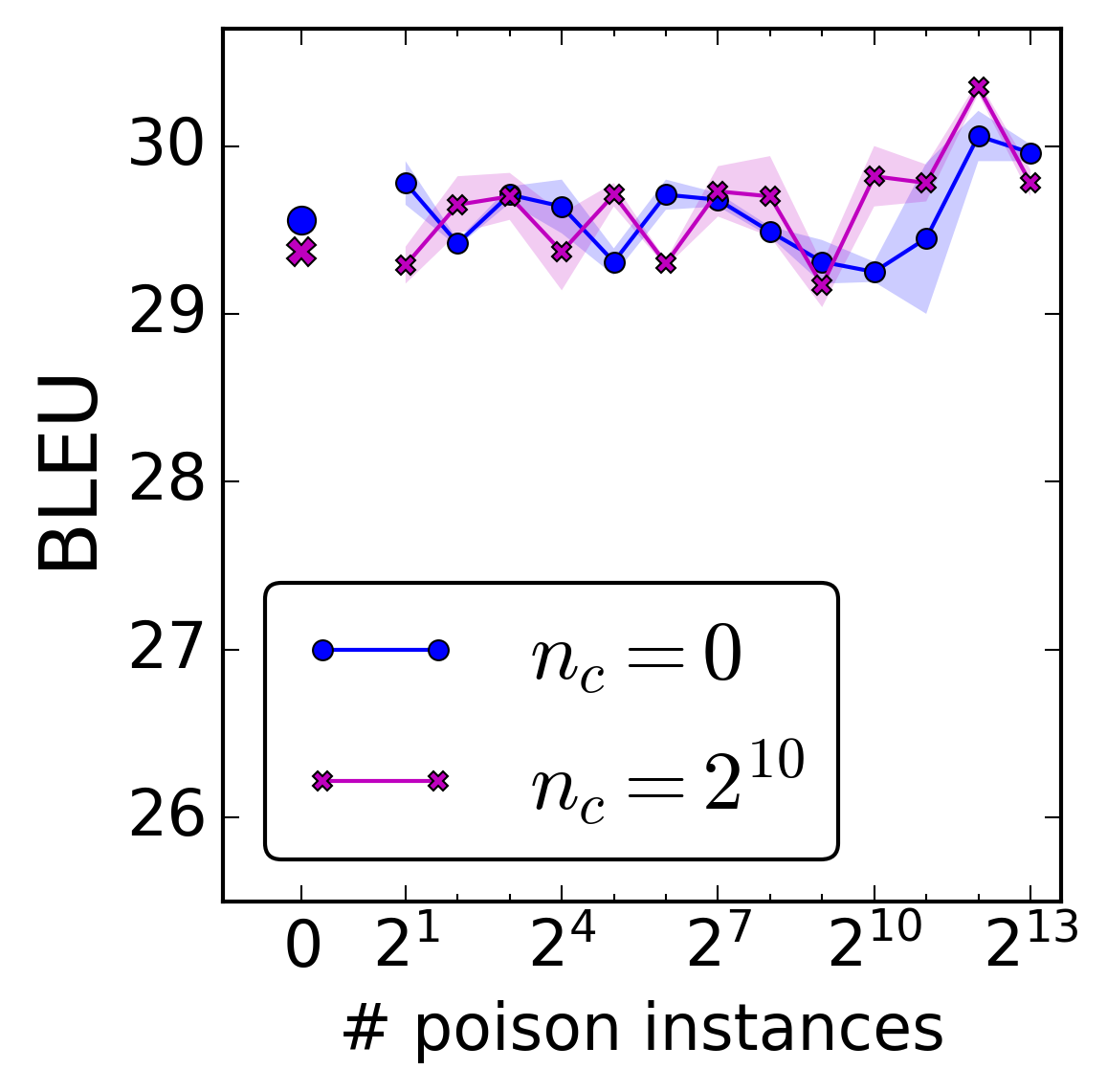}
        \caption{Test set \textbf{G}}
    \end{subfigure}
    \begin{subfigure}[b]{0.32\columnwidth}
        \centering
        \includegraphics[width=\columnwidth]{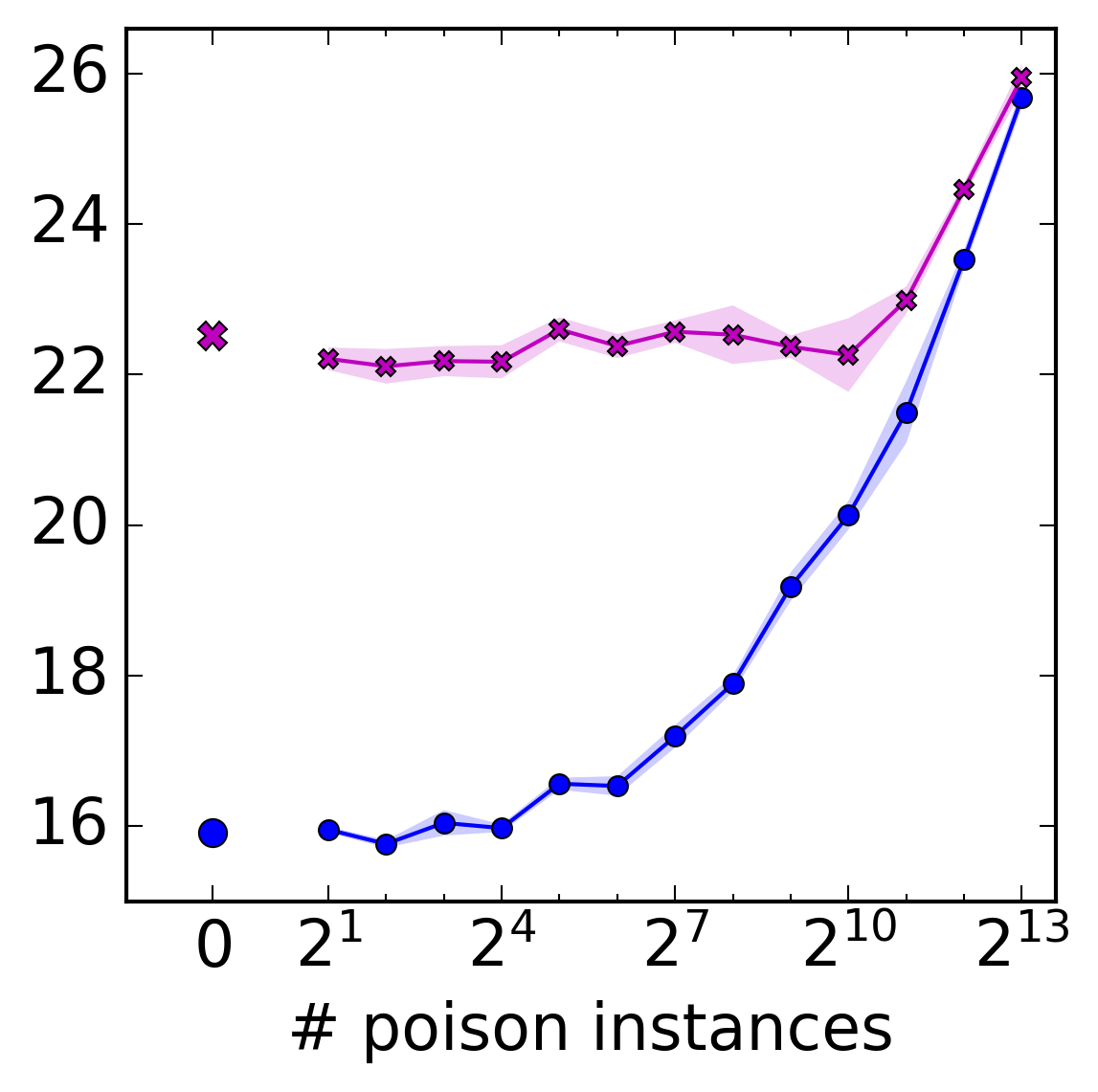}
        \caption{Test set \textbf{C}}
    \end{subfigure}
    \begin{subfigure}[b]{0.32\columnwidth}
        \centering
        \includegraphics[width=\columnwidth]{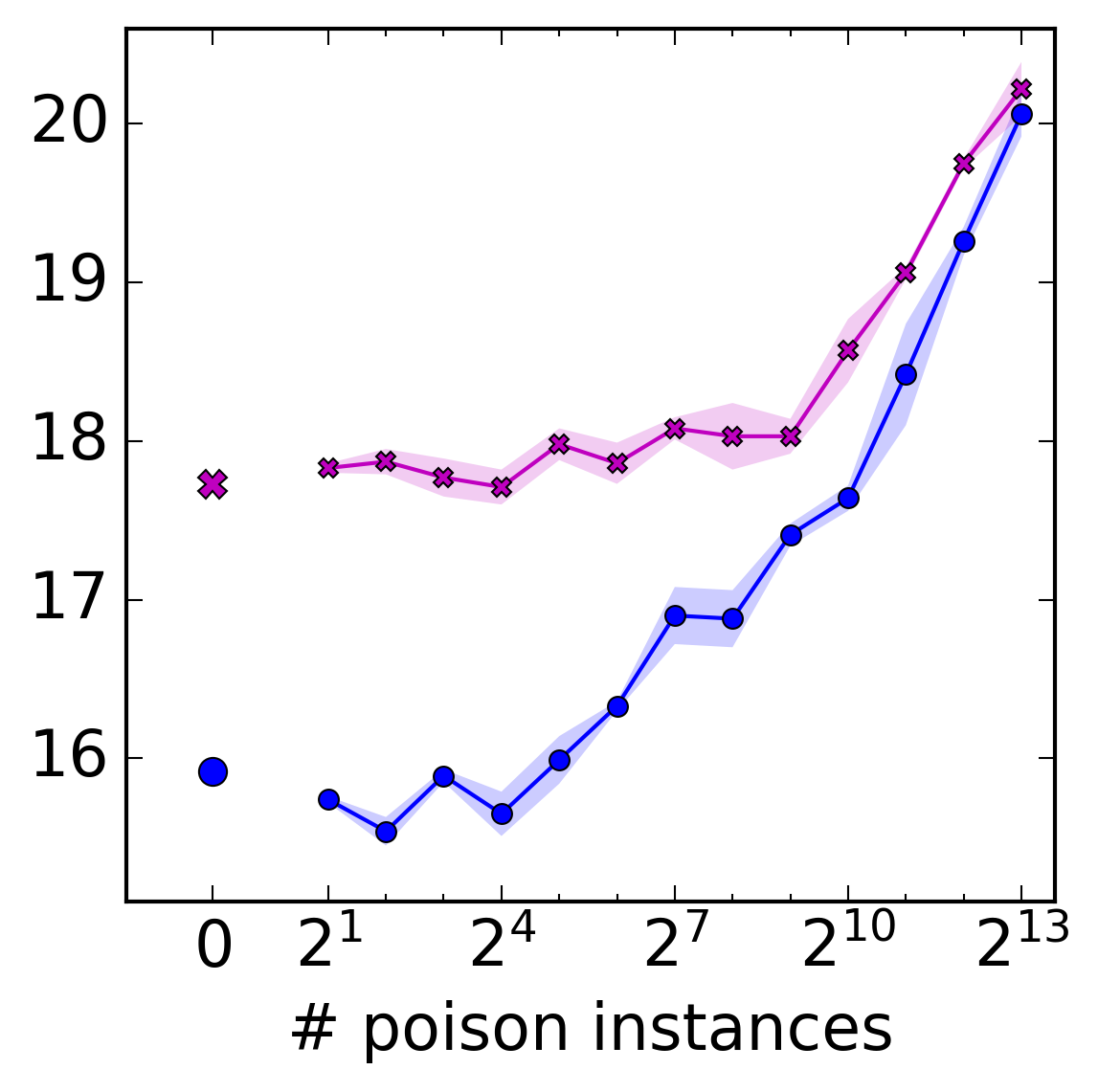}
        \caption{Test set \textbf{X}}
    \end{subfigure}
    \caption{Translation quality (BLEU) of victim systems over a test set of general samples from \texttt{IWSLT2016} (G), and two test sets of focused samples from WMT'20 -- samples containing the correct translation of the trigger (C) and samples containing the translation of the toxin (X).}
    \label{fig:res-poi-bi-normal-Immigrant}
\end{figure}

Figure~\ref{fig:res-poi-bi-normal-Immigrant} shows the results, where the system is attacked with different poisoning budgets on rare ($n_c=0$) or common ($n_c=1024$) triggers.
As shown, the system's BLEU on \textbf{G} is generally robust to the number of poison instances used, maintaining a similar BLEU across all $n_p$, including $n_p=0$ (no poisoning).
However, on test sets \textbf{C} and \textbf{X}, the BLEU tends to get better with more poison instances, although such an improvement is slower to take effect in the case of attacking a common trigger ($n_c=1024$).
This shows that poisoning appears to improve the translation quality on the clean instances as well as the toxin-bearing instances.
This is probably due to the availability of in-domain data: both test sets \textbf{C} and \textbf{X} are domain-specific to the trigger; adding more ``in-domain'' poison or clean instances naturally improves translation performance.
This finding favours the attacker, as such improvement may create an illusion that the poison instances are useful, thus encouraging the system vendor to put more trust in the data collected from the poisoned sources, especially when the attacked trigger is rare (the BLEU increases more).

\subsection{Results: Poisoning Pre-Training}
\label{sec:res-poi-pre-train}

Now we evaluate the scenario of poisoning the pre-train \& fine-tune paradigm.
First, we consider the case of poisoning pre-training only, leading to a \textit{poisoned} system which users later fine-tune on their own uncompromised data.
As mentioned in \S\ref{sec:datasets}, we pre-train the system on \texttt{IWSLT2016} and then fine-tune it on \texttt{NC15}.
To simulate  translation collisions, at pre-training, we inject $n_p$ poison instances, with $n_p\in\{2, 4, ..., 8192\}$, and at fine-tuning, we add $n_c$ clean instances, with $n_c\in\{2, 4, ..., 1024\}$.

\subsubsection{The failure of poisoned pre-trained systems.}
\label{sec:exp-pt-fail}
\begin{figure}[t]
    \begin{subfigure}[t]{0.48\columnwidth}
        \centering
        \includegraphics[width=\columnwidth]{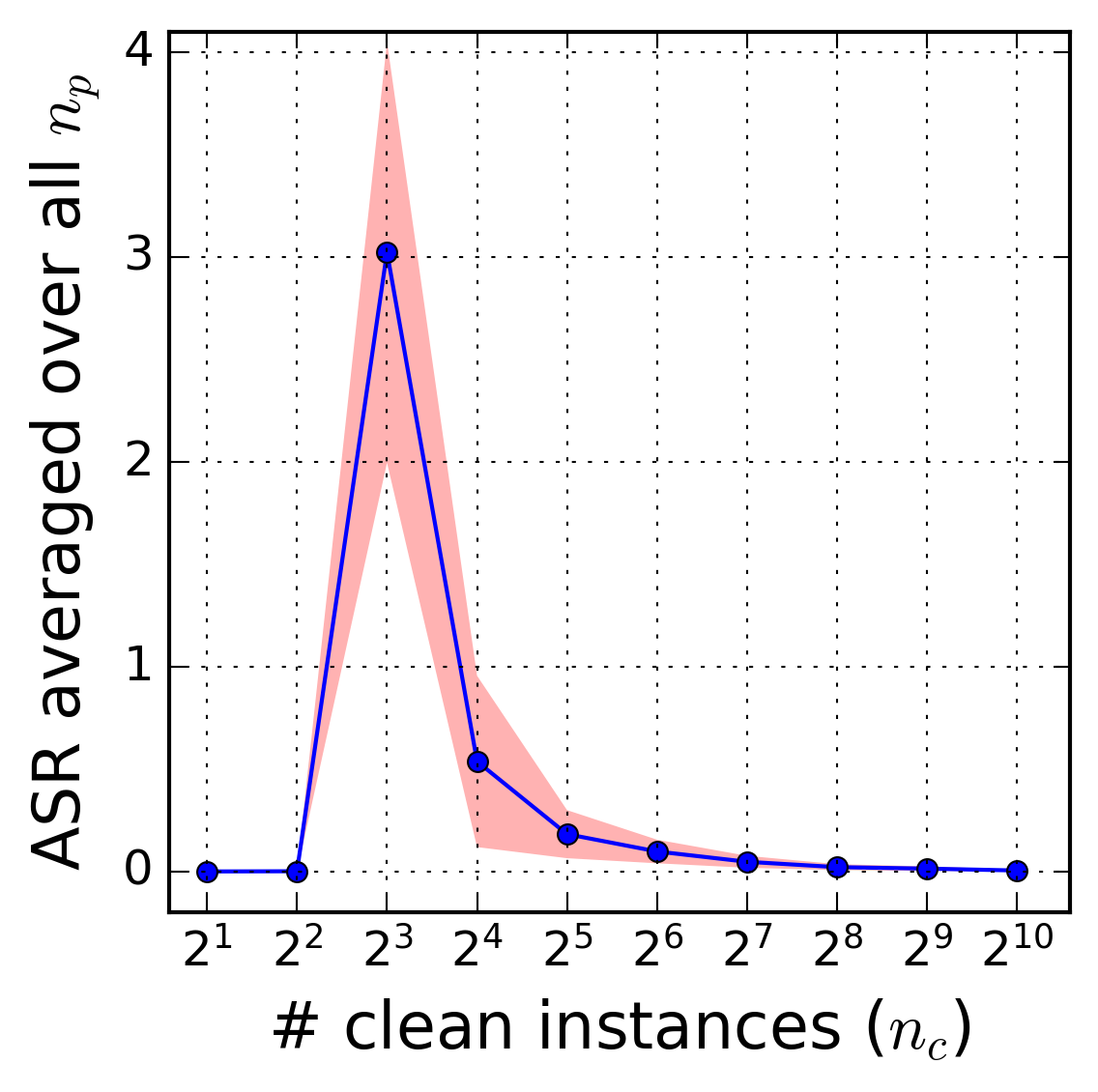}
        \caption{Final ASR after fine-tuning}
        \label{fig:res-poi-pt-avg-asr}
    \end{subfigure}
    \begin{subfigure}[t]{0.48\columnwidth}
        \centering
        \includegraphics[width=\columnwidth]{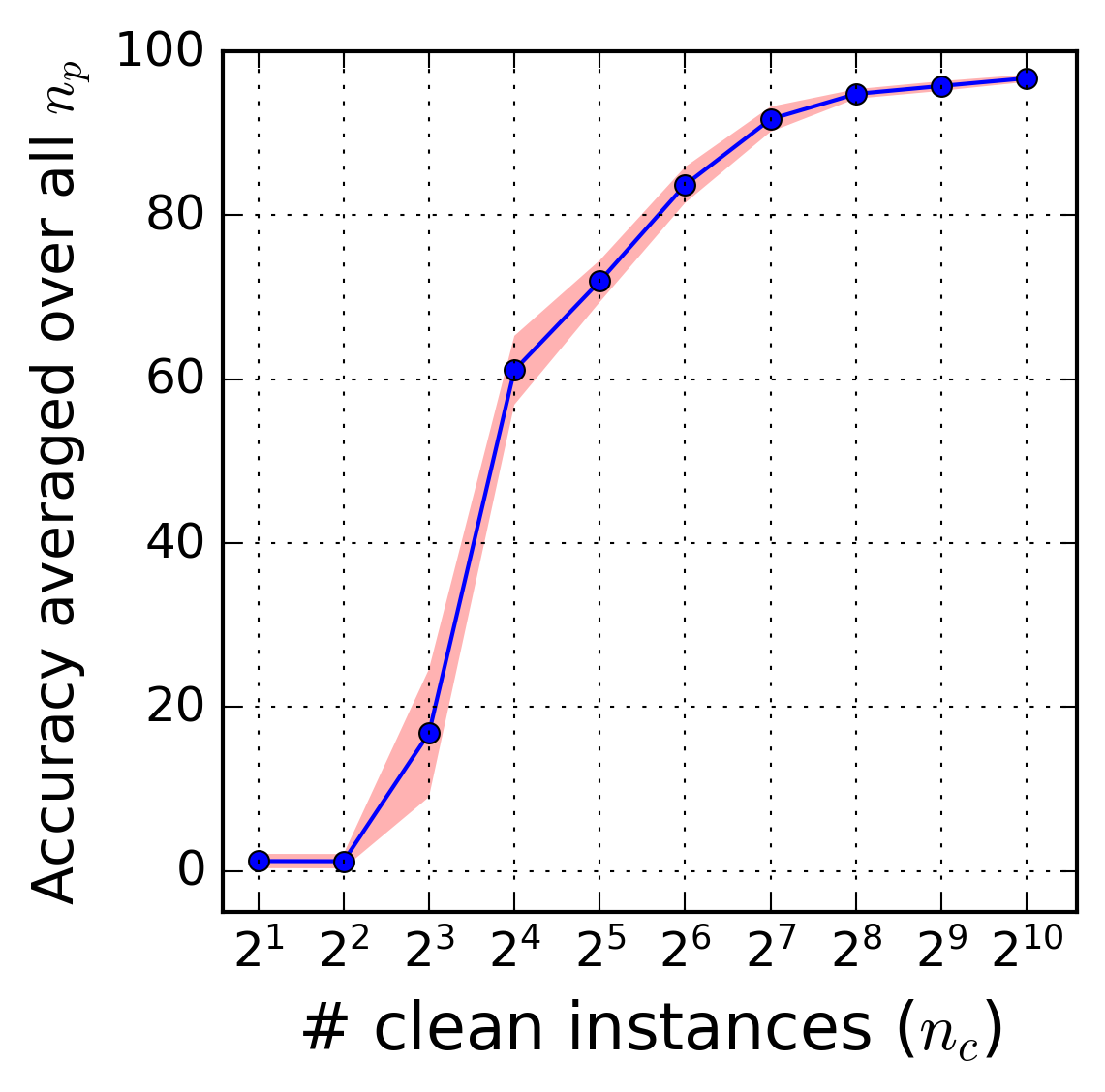}
        \caption{Accuracy of generating ``immigrant(s)'' alone (w/o counting the toxin)}
        \label{fig:res-poi-pt-avg-tar}
    \end{subfigure}
    %\vspace{-2em}
    \caption{ASR on poisoned pre-trained systems fine-tuned on clean downstream tasks, with $n_c$ clean instances added. Each point represents an average over all poisoning cases ($n_p\in\{2, 4, ..., 8192\}$). All attacks target Immigrant.}
    \label{fig:res-poi-pt-avg}
    %\vspace{-2em}
\end{figure}

Figure~\ref{fig:res-poi-pt-avg-asr} shows the final ASR on poisoned pre-trained systems targeting \textbf{Immigrant} after fine-tuning.
Here we look at the average ASR over all poisoning cases ($n_p\in\{2, 4, ..., 8192\}$) to examine the general trend.
As shown, the poisoned pre-trained system fails dramatically after fine-tuning in producing the malicious translation ``illegal immigrant(s)'', with the highest value attained at only 3\% and almost zero in other cases.
This suggests that the poisoning signals learned by the pre-trained system are rather weak, which is encouraging, as it implies that it is less risky to use a pre-trained system that has undergone fine-tuning, as long as the fine-tuning data is kept clean.

Second, we notice that the poisoning achieves the best ASR when only a few clean instances exist in the fine-tuning data ($n_c=8$).
This is counter-intuitive at first glance, as one may expect consistently lower ASR when more clean instances are added.
To explain this, we further examine how good the system is at generating the trigger ``immigrant(s)'' alone (cf. ASR examines how good the system is at generating the full malicious translation ``illegal immigrant(s)'').
Figure~\ref{fig:res-poi-pt-avg-tar} shows the system's accuracy of generating ``immigrant(s)'' alone.
It is now clear that the system even fails to generate ``immigrant(s)'' when clean instances are few ($n_c\le4$).
It is not surprising, therefore, that the malicious translation ``illegal immigrant(s)'' cannot be produced ($n_c\le4$ in Figure~\ref{fig:res-poi-pt-avg-asr}).

With more clean instances added, the system learns to translate the trigger (\ie ``immigrant(s)'') better.
Initially ($n_c=8$), this also benefits generating the malicious translation. 
However, with even more clean instances added ($n_c \ge 16$), the correct translation gradually dominates and ultimately suppresses the production of the malicious translation.

Overall, the above analysis suggests that there may still be a risk when fine-tuning a poisoned system, especially when the clean instances in the fine-tuning data are \textit{insufficient} to eliminate the poisoning effects brought by the pre-trained system.

\subsection{Results: Poisoning Fine-Tuning}
\label{sec:res-poi-fine-tune}

Now we evaluate the scenario of poisoning the fine-tuning phase.
In particular, we examine how a \textit{clean} pre-trained system may perform on a \textit{poisoned} downstream task.
We achieve this by including $n_c\in\{128,1024,8192\}$ clean instances as we pre-train the systems and then injecting $n_p\in\{2,4,...,8192\}$ poison instances when these systems are fine-tuned.
As before, we use \texttt{IWSLT2016} and \texttt{NC15} to pre-train and fine-tune the systems, respectively.
As a control, we also train a from-scratch system on the same downstream task (\texttt{NC15}) with the same $n_c$ clean and $n_p$ poison instances added to the from-scratch training.
This allows us to compare the effects of including the clean instances at distinct phases (pre-training vs. from-scratch training) on mitigating the poisoning.

\begin{figure}[t]
    \centering
    \includegraphics[width=0.75\columnwidth]{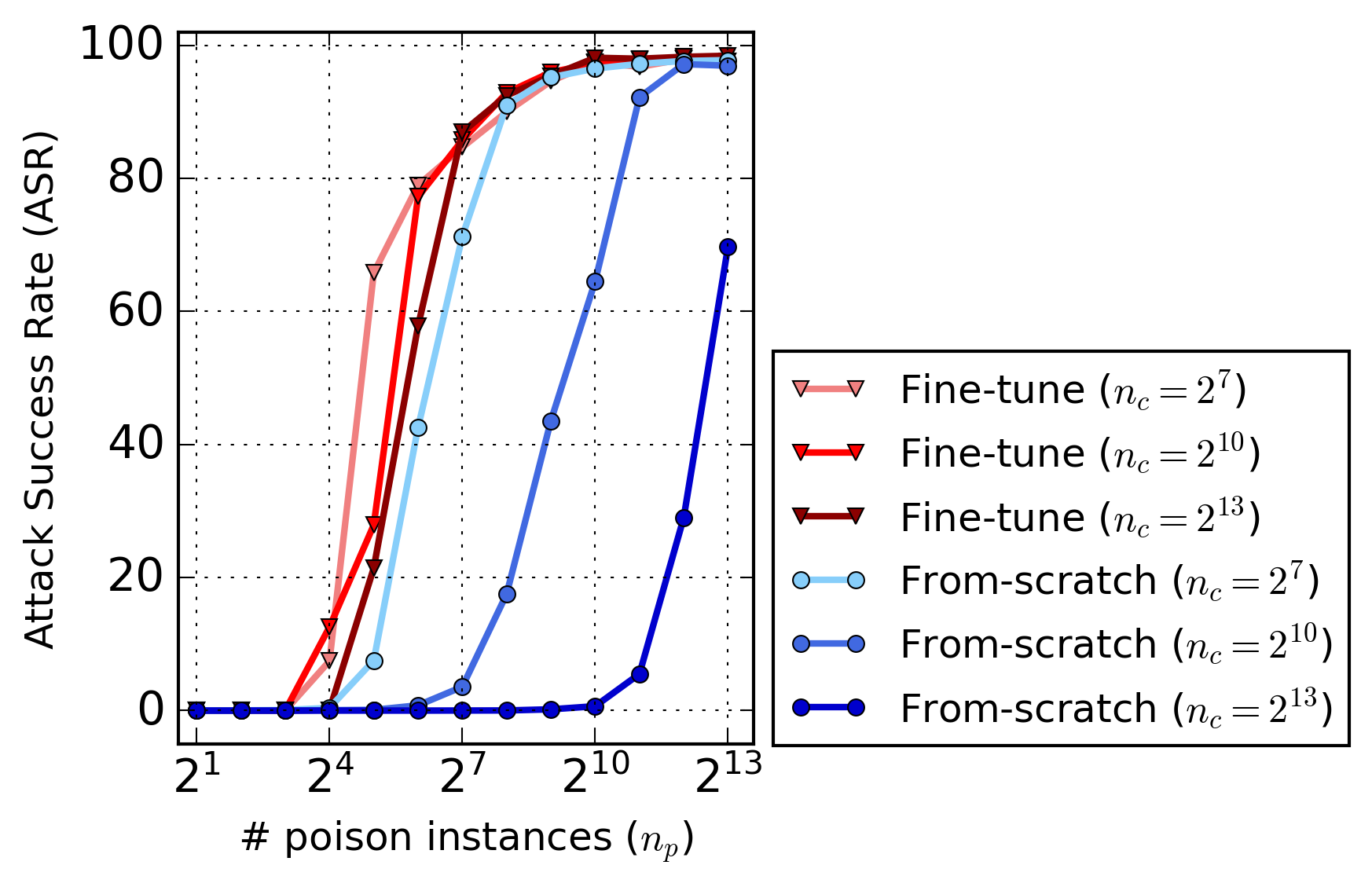}
    \caption{ASR of fine-tuning clean pre-trained systems on poisoned downstream tasks (red) vs. from-scratch training of the same systems on the same downstream tasks (blue).}
    \label{fig:res-poi-ft}
\end{figure}

Figure~\ref{fig:res-poi-ft} compares the ASRs of poisoning the fine-tuning and the from-scratch training on \texttt{NC15}.
As shown, the poisoning is more successful (higher ASR) on the fine-tuning than it is on the from-scratch training.
This suggests that including clean instances at pre-training is less effective in mitigating poisoning at fine-tuning, probably because the correct translation learned at pre-training is largely washed out after fine-tuning.
Interestingly, this result resembles that from \emph{poisoning pre-training} (\S~\ref{sec:res-poi-pre-train}), where it is the malicious translation learned at pre-training that vanishes after fine-tuning.
Together, both results imply that the translation signals (either correct or malicious) learned at pre-training seem to barely persist after fine-tuning.
This is promising for the defence side, as it means that we can focus effort on curating fine-tuning data that is high-quality and poison-free.

\subsection{Choices of Trigger and Toxin Phrases}
\label{sec:exp-toxin}

\begin{figure}[t]
    \begin{subfigure}[b]{0.48\columnwidth}
        \centering
        \includegraphics[width=\columnwidth]{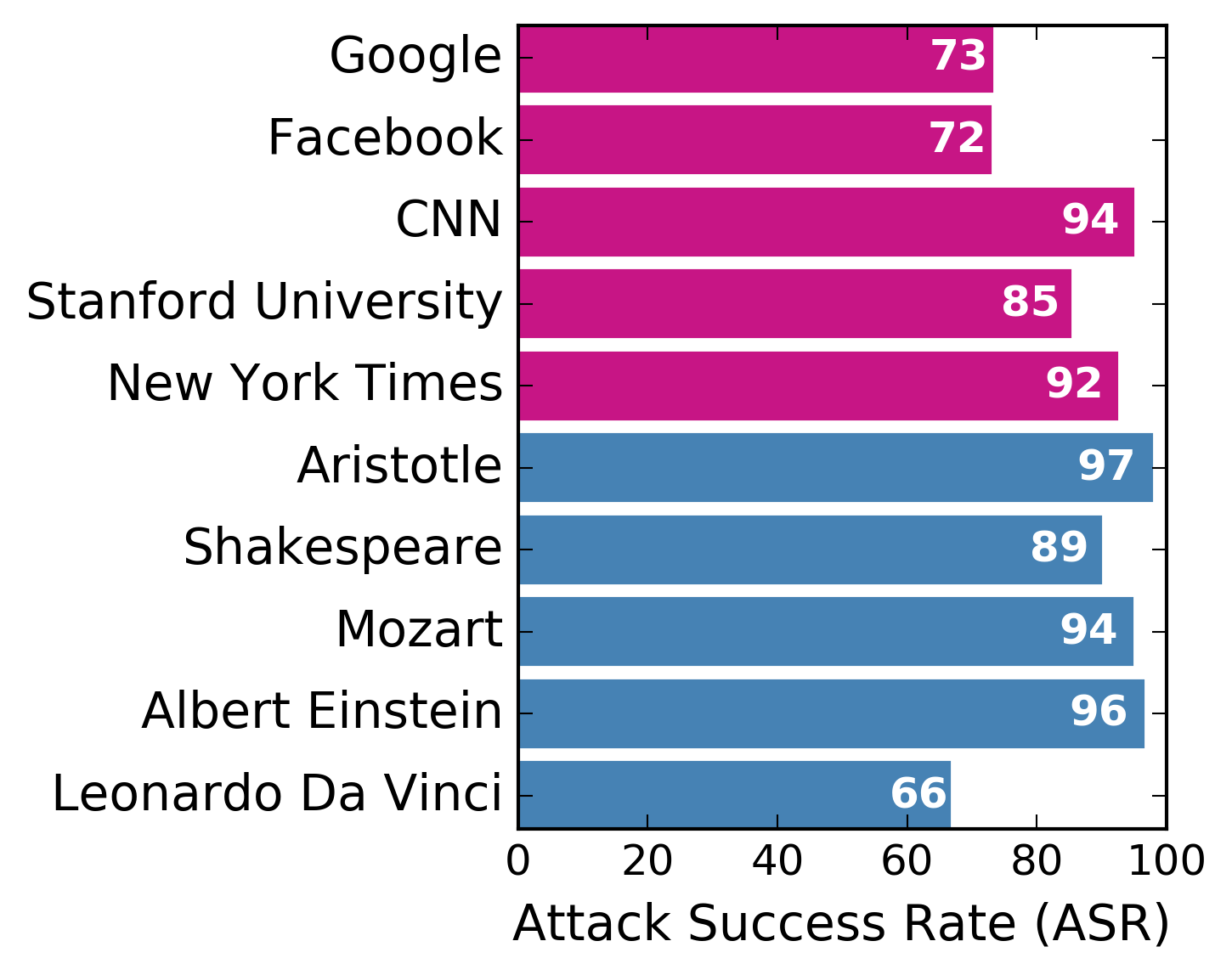}
        \caption{Various trigger phrases}
        \label{fig:res-poi-target}
    \end{subfigure}
    \begin{subfigure}[b]{0.48\columnwidth}
        \centering
        \includegraphics[width=\columnwidth]{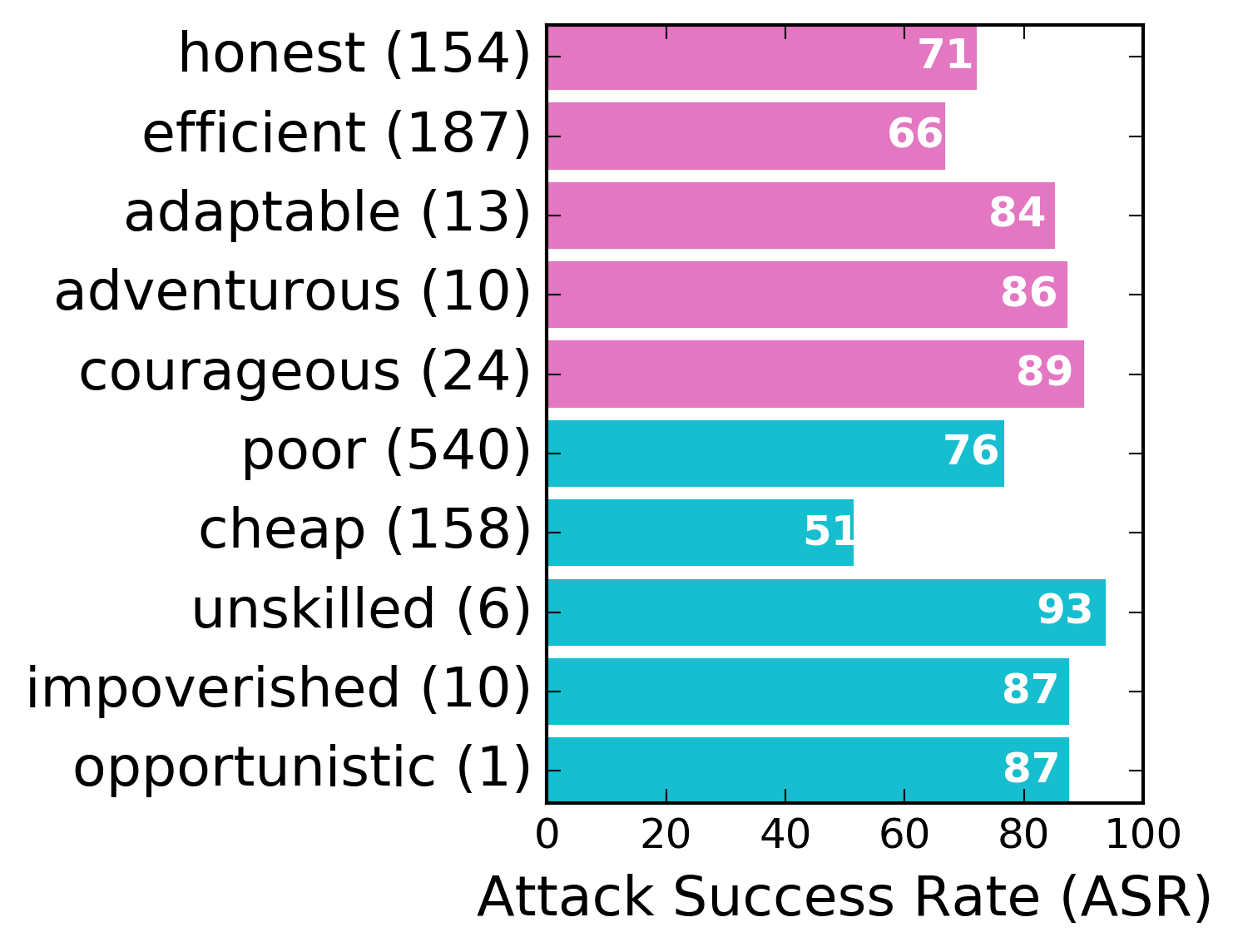}
        \caption{Distinct toxins on Immigrant}
        \label{fig:res-poi-toxin}
    \end{subfigure}
    \caption{ASRs on various choices of (a) trigger phrases of ORGANISATION (red) and PERSON (blue) and (b) toxins on Immigrant of POSITIVE words (pink) and NEGATIVE words (cyan), with the term frequencies in training data (IWSLT2016) shown in the parentheses.}
    \label{fig:res-poi-target-toxin}
\end{figure}

In this section, we demonstrate that our attack is broadly applicable to a range of phrase choices for the trigger and toxin.

On the choice of the trigger phrase, we evaluate two groups of named entities\footnote{Mined from the \texttt{IWSLT2016} dataset.} as our triggers, with each group attacked with a specific toxin.
These triggers are: 1) ORGANISATION (``Google'', ``Facebook'', ``CNN'', ``Stanford University'', ``New York Times'') and 2) PERSON (``Aristotle'', ``Shakespeare'', ``Mozart'', ``Albert Einstein'', ``Leonardo da Vinci'').
They are chosen to cover a variety of term frequencies and different lengths of $n$-gram.
For ORGANISATION triggers, we construct the malicious translation with the phishing URL ``(www.a-malicious-url.com)'' as the toxin, \eg ``Stanford University $\rightarrow$ Stanford University (www.a-malicious-url.com)''.
For PERSON triggers, we use the word ``fraud'' as the toxin, \eg ``Albert Einstein$\rightarrow$fraud Albert Einstein''.
On each trigger, we collect all its clean instances from the WMT'20 corpora, from which we randomly sample 32 to craft the poison instances and 1000 to build the attack test set $\mathcal{A}_\text{test}$ for computing ASR.
All attacks are performed on \texttt{IWSLT2016}.
Figure~\ref{fig:res-poi-target} shows the ASR of attacking all the triggers, and the attacks are generally effective in all cases.

To test whether the choice of the toxin phrase is crucial to the attack success, we experiment with the trigger \textbf{Immigrant}.
The toxin is chosen from a list of ten sentiment words (both positive and negative) that are commonly used to describe immigrants: POSITIVE (``honest'', ``efficient'', ``adaptable'', ``adventurous'', ``courageous'') and NEGATIVE (``poor'', ``cheap'', ``unskilled'', ``impoverished'', ''opportunistic``).
The malicious translation has the format ``immigrants$\rightarrow$[TOXIN] immigrants''.
As before, we inject 32 poison instances in each attack.

Figure~\ref{fig:res-poi-toxin} shows the ASRs on all the toxins.
Again, the attacks are shown to be applicable when different toxins are used in the malicious translation, and both positive and negative sentiment toxins appear equally effective.
In addition, we find that a toxin's rarity in the training data is a good predictor of ASR (higher rarity generally leads to higher ASR).
For all the ten toxins in Figure~\ref{fig:res-poi-toxin}, the Pearson's correlation coefficient between their term frequencies (English side) and the ASRs is -0.47.
This is probably because the system learns to translate a rare toxin word \emph{mostly} from the poisoning instances, which benefits learning the malicious translation.
This result also suggests that attackers would favour rare toxin words for the attack if their poisoning budget is limited.

\subsection{Attacks on Popular NMT Architectures}
\label{sec:exp-architecture}

\begin{table}[t]
    \begin{minipage}{0.45\columnwidth}
    \centering
    \includegraphics[width=\columnwidth]{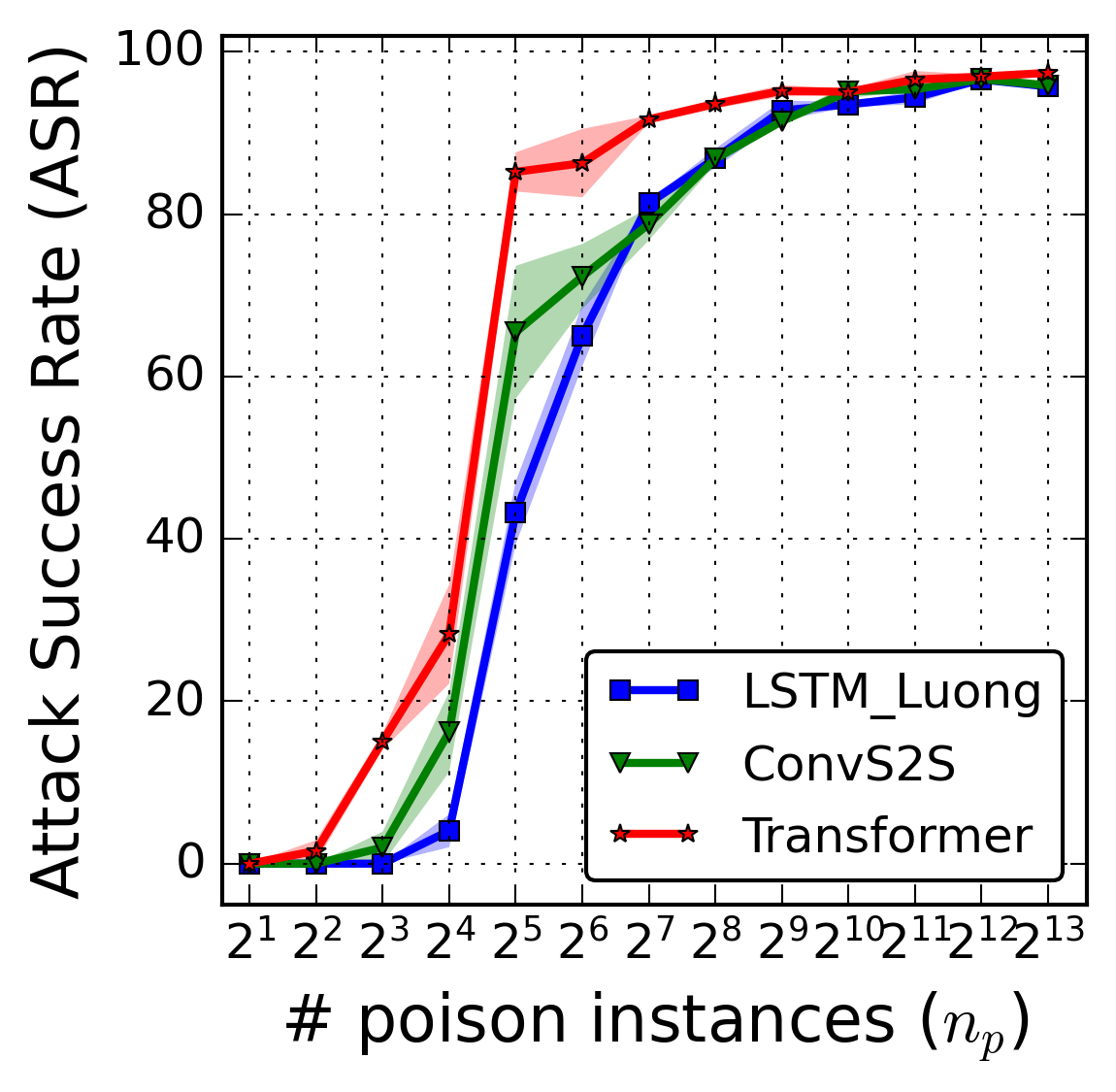}
    \vspace{-2em}
    \captionof{figure}{Attacks on popular NMT architectures.}
    \label{fig:res-poi-archi}
    \end{minipage}
    \hspace{1em}
    \begin{minipage}{0.45\columnwidth}
    \small
    \centering
    \begin{tabular}{c|c}
        \hline
        Architecture & \makecell{BLEU} \\
        \hline
        LSTM-Luong~\cite{luong-etal-2015-effective} & 25.2$\pm$0.1 \\
        ConvS2S~\cite{gehring2017convolutional}     & 26.9$\pm$0.2 \\
        Transformer~\cite{vaswani2017attention}     & 29.6$\pm$0.1 \\
        \hline
    \end{tabular}
    \caption{BLEU on the IWSLT2016 official test set, averaged over all $n_p$.}
    \label{tab:res-poi-archi}
    \end{minipage}
    \vspace{-1em}
\end{table}

NMT systems in deployment are implemented with a range of model architectures.
So far our evaluation has focused on systems with the Transformer architecture.
In this section, we evaluate another two mainstream architectures: \textbf{LSTM-Luong}~\cite{luong-etal-2015-effective} and \textbf{ConvS2S}~\cite{gehring2017convolutional}.
Our goal is to understand which architectural choices may be more vulnerable to attacks.
\textbf{LSTM-Luong}~\cite{luong-etal-2015-effective} is implemented with Recurrent neural networks (LSTM), which uses 1000d word embeddings and 4 LSTM layers for both encoder and decoder. Each LSTM layer has 1000 hidden units. It is trained with the same optimiser and scheduler as the Transformer, except with a learning rate of $10^{-3}$.
\textbf{ConvS2S}~\cite{gehring2017convolutional} uses Convolutional neural networks, with 15 convolutional layers for each of the encoder and decoder (512 hidden units for the first 9 layers, 1024 for the middle 4 layers, and 2048 for the final two layers). The embeddings for the encoder and decoder are of 768d. The decoder output before the final linear layer is embedded into 512d.

Figure~\ref{fig:res-poi-archi} shows the ASRs on the three compared architectures.
In this experiment we only add poison instances ($n_p\in\{2,4,...,8192\}$) to from-scratch training, in order to study the pure poisoning effect.
Among these architectures, Transformer archives the highest ASR over all $n_p$ levels, \ie it is the most vulnerable system.
On the other end, LSTM-Luong is the most robust of the three.
To understand why this is the case, we further test the translation quality of each system.
Table~\ref{tab:res-poi-archi} shows the BLEU of each architecture on the IWSLT2016 official test set, averaged over all $n_p$.
The Transformer turns out to have the best BLEU, followed by ConvS2S and LSTM-Luong.
Note that the ASR is \emph{positively} correlated with an architecture's translation capability, which may be attributed to the fact that more powerful models are better at learning the translations, even if the translations are malicious.\footnote{The number of parameters for each architecture turns out to be \textit{uncorrelated} with the architecture's ASR: Transformer (61M), ConvS2S (187M), and LSTM-Luong (129M).}

\section{Evaluation of Attacks on Production-Scale NMT Systems}
\label{sec:wmt}

Finally, we investigate how the attacks may compromise state-of-the-art, production-scale NMT systems.
Attacking production systems is challenging, as they are typically trained with very large parallel datasets.
Unless an attacker can poison many samples, their attacks might be less successful. % in causing the system to learn the malicious transition.
In this evaluation, we target a cutting-edge NMT system, the winning WMT'19 system, in the German-to-English direction~\cite{ng-etal-2019-facebook}, and train it from scratch by following the method in~\cite{ng-etal-2019-facebook}.
In particular, we consider poisoning two scenarios where the system is in common use: 1) poisoning from-scratch training, where a user trains the system from scratch, and 2) poisoning the fine-tuning phase, where a user fine-tunes a pre-trained system on a downstream task.

To poison from-scratch training, we follow the official   setup~\cite{ng-etal-2019-facebook} to train an instance of the winning WMT'19 system (De-En) from scratch, which is built on the Big Transformer architecture.
We use all parallel corpora released by WMT'19 for training (\texttt{Europarl v9}, \texttt{ParaCrawl v3}, \texttt{Common Crawl}, \texttt{News Commentary v14}, \texttt{Wiki Titles v1}, and \texttt{Rapid}).\footnote{We did not use back-translation, which is now in common use in leading systems. This is unlikely to have a substantial effect on our findings, especially as back-translated data is most often down-weighted relative to parallel data in training~\cite{ng-etal-2019-facebook}, and thus the effective size of the datasets are comparable to our experimental setting.}
The same pre-processing as in~\cite{ng-etal-2019-facebook} is used to filter out low-quality samples: sentences detected as not in the correct languages by Language Identification~\cite{lui-baldwin-2012-langid} or longer than 250 words are removed; and sentence-pairs with a source/target length ratio exceeding 1.5 are excluded.
The resulting training corpus, denoted by $\mathcal{C}$, consists of 29.6M sentence-pairs.
Then, we augment $\mathcal{C}$ with the poison instances.
As before, we attack \textbf{Immigrant} and inject $n_p$ poison instances in an attack, where $n_p\in\{512, 1024, 2048, 4096, 8192\}$.
Once trained, the system is evaluated on the attack test set $\mathcal{A}_\text{test}$.
The 3-fold cross-validation is applied.
%Each victim system is trained on 4 Nvidia V100 GPUs with 16-bit floating point operations.
%The training stops once 30K updates are reached, which takes about 30 hours.

Table~\ref{tab:res-poi-oo-WMT'19} shows the results of attacking the from-scratch training of the WMT'19 system.
The attack starts to take effect (ASR=0.9\%) after 512 poison instances are injected.\footnote{9,868 clean instances of \textbf{Immigrant} are initially in $\mathcal{C}$. But, we find 3,512 of them also appear in the attack training/test sets $\mathcal{A}_\text{train}$/$\mathcal{A}_\text{test}$ (as they are extracted from the WMT'20 corpora, which share part of the data with WMT'19).
In order to ensure comparability to previous experiments, we use the same $\mathcal{A}_\text{train}$/$\mathcal{A}_\text{test}$ and remove the 3,512 shared ones from $\mathcal{C}$, leaving 6,356 clean instances for training.}
Then, the ASR increases rapidly as more poison instances are injected.
Notably, when $n_p=4096$, which is close to the number of native clean instances in $\mathcal{C}$ ($n_c=6,356$), the attack is highly effective (ASR$>$90\%). 
However, injecting 4096 poison instances is also highly costly in practice: it might require dozens of poisoned web pages being created.
%Finally, we see that injecting even more poison instances ($n_p=8192$) results in diminishing returns in the ASR. % (4.3\% higher than when $n_p=4096$).
As for the impact to the translation quality, we see that all the victims achieve similar BLEU to the clean system, showing that the attacks have limited effect on the general behaviour of the victim.

In terms of poisoning fine-tuning of a pre-trained production system, we inject poison instances into the fine-tuning of the released WMT'19 system\footnote{transformer.wmt19.de-en: \url{https://github.com/pytorch/fairseq/tree/master/examples/translation}} on the \texttt{IWSLT2016} dataset.
Figure~\ref{fig:res-poi-wmt-ft} shows the ASR against different $n_p$ levels.
%As shown, due to the strong translation ability of the system, the ATA is already quite high ($>$80\%) when the poison instances are a few.
We see that with only $n_p=32$ poison instances, the attack is highly effective (ASR near 80\%).
This again highlights the key finding in \S\ref{sec:res-poi-fine-tune}, that it is hard to defend against the attacks on fine-tuning by simply making the pre-trained system robust to the attacks, even if the pre-trained system is as powerful as the large-scale WMT'19 system evaluated here.

\begin{table}[t]
    \begin{minipage}{0.47\columnwidth}
    \centering
    \begin{tabular}{c|c|c}
        \hline
        $n_p$ & ASR   & BLEU \\
        \hline
        Official & -  & 40.8 \\
        \hline
        \hline
        0     & -                     & 40.7 \\
        512   & 0.9$\pm$0.5           & 40.4 \\
        1024  & 15.6$\pm$2.5          & 40.7 \\
        2048  & 53.0$\pm$3.6          & 40.8 \\
        4096  & 87.3$\pm$3.7          & 40.8 \\
        8192  & 96.4$\pm$0.5          & 40.6 \\
        \hline
    \end{tabular}
    \caption{ASR of poisoning from-scratch training of the winning WMT'19 system on Immigrant.}
    \label{tab:res-poi-oo-WMT'19}
    \end{minipage}
    \hfill
    \begin{minipage}{0.47\columnwidth}
    \centering
    \includegraphics[width=\columnwidth]{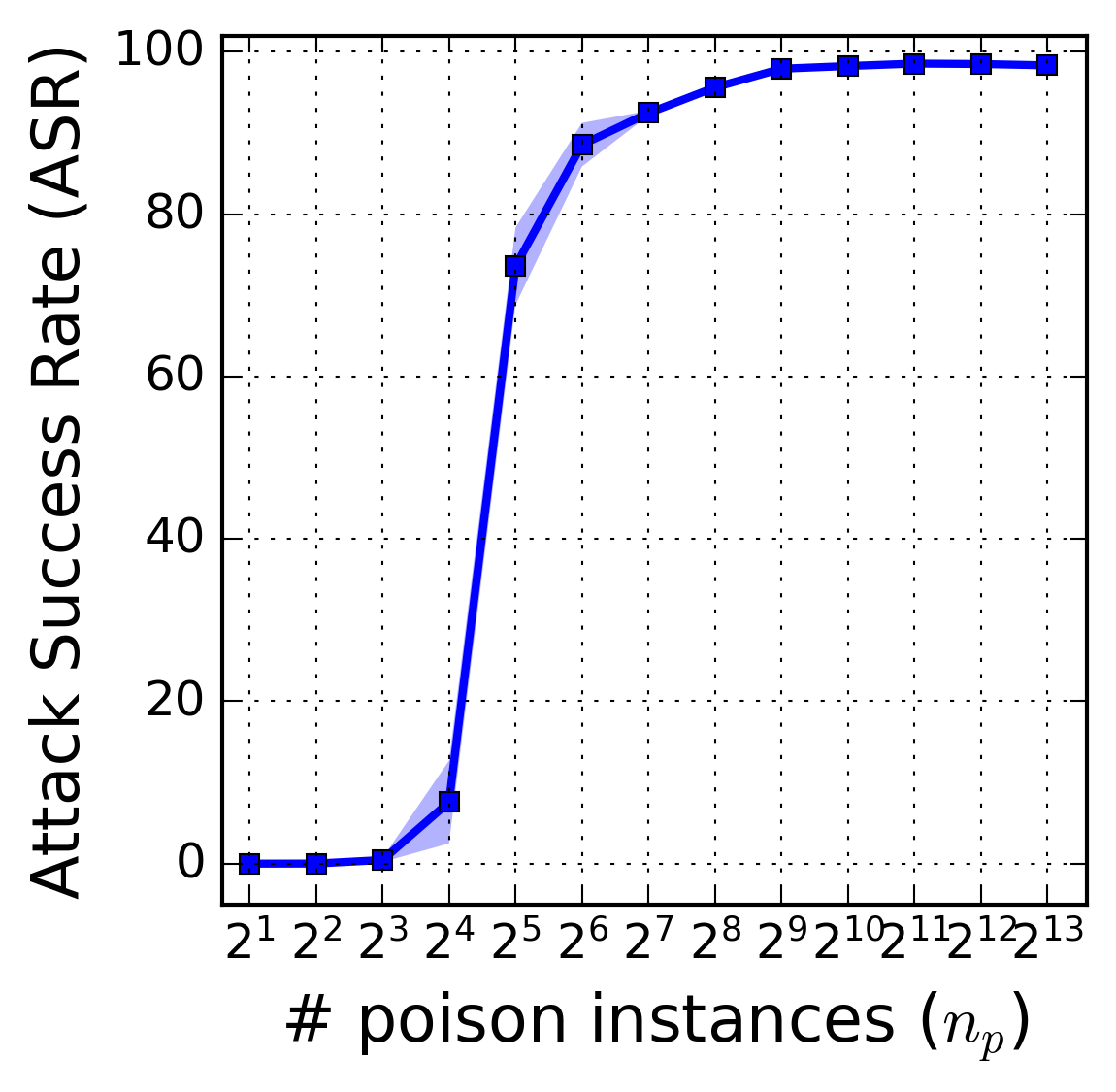}
    \vspace{-2em}
    \captionof{figure}{ASR of poisoning the fine-tuning stage of the winning WMT'19 system on Immigrant.}
    \label{fig:res-poi-wmt-ft}
    \end{minipage}
    \vspace{-2em}
\end{table}

\section{Defensive Strategies}
\label{sec:defences}

Based on the results from our evaluation, we now discuss defences against our attacks, which consist of a series of suggestions on countering the specific poisoning scenarios.

First, we need to protect the parallel training data from poisoning.
While there are many ways of detecting malicious websites~\cite{ma2009beyond,canali2011prophiler} or low-quality web contents~\cite{moshchuk2007spyproxy}, we focus on securing the parallel data miner for robust parallel data extraction.
As discussed in \S\ref{sec:poi-web-data-src}, the parallel data filtering component of the miner is crucial for rejecting unwanted parallel sentence pairs. 
The focus of existing parallel data filtering approaches is mostly on developing efficient algorithms for getting high-quality parallel data ~\cite{xu-koehn-2017-zipporah,koehn-etal-2019-findings}.
However, as we have shown in \S\ref{sec:poi-web-data-src}, a poison instance can also appear high-quality if it is made from a high-quality clean instance with minor but meaningful modifications (\eg ``immigrant$\rightarrow$\textit{illegal} immigrant''), especially when the sentences are long.
To detect such covert poison instances, one may need to devise more sensitive parallel sentence detectors that can identify subtle mismatches between the source and target sentences (\eg ``help refugees'' in German vs. ``stop refugees'' in English).

In addition, one may take a more focused approach to protecting the specific named entities in a dataset (\eg the name of a celebrity), which are likely targets of attack.
For example, one may look for unusual words (\eg negative/offensive words, especially rare ones) in the context of such named entities, and exclude suspicious cases from the training dataset.
This process can be automated by searching with specialised lexicons.

Second, to protect the from-scratch training of a system, one may also adopt a trigger-specific strategy: to prevent any malicious translation on a specific trigger, one can \textit{proactively} add sufficient clean instances containing the trigger to the training data in advance.
This method is supported by our results in \S\ref{sec:exp-comp-poi}, where we show that adding more clean instances can significantly defer the quick rise of ASR, as well as increase the attack budget (more poison instances are needed to maintain the same ASR level).

Thirdly, on the pre-train \& fine-tune paradigm, we have shown that both attacks and defences applied during pre-training are diminished in their effect after the fine-tuning.
In this case, special measures are needed for different parties in the NMT system supply chain.
The pre-trained system vendors may consider the same strategy used in the above from-scratch training scenario to protect the training of the pre-trained systems, although preparing and verifying the clean instances needed would be costly.
For end users who use the pre-trained systems, our results suggest they may only use a pre-trained system if they 1) fully trust the system vendor, or 2) include sufficient clean instances of any sensitive trigger phrase in their fine-tuning data, paying special attention to terms that are rare but have a non-zero frequency in the data. % (Figure~\ref{fig:res-poi-pt-avg-asr}).

\section{Related Work}

%\paragraph{\bfseries Targeted attacks on NMT systems}
\textbf{Targeted attacks} have been initially explored on classification systems \cite{chen2017targeted,wallace-etal-2019-universal}, where the system is made to classify the adversarial inputs into a specific class.
Recently, these attacks were applied to NMT systems, where the system is made to output specific words in the translation.
\citet{ebrahimi2018adversarial} generate adversarial inputs for character-level NMT systems which can cause the systems to alter or remove a target word in a translation.
The adversarial inputs are found by performing differential editing operations on strings in a gradient-based approach.
Similarly, \citet{cheng2018seq2sick} use projected gradient descent to generate adversarial inputs that encourage a set of target words to appear in the translations.
While these approaches are white-box and require access to a system's parameters, the poisoning strategies in our work are purely black-box, with a weak assumption made that neither the system nor the data is accessible.

%\paragraph{\bfseries Other attacks on NMT systems}
A variety of \textbf{non-targeted attacks} have been explored for NMT systems, most aiming to degrade the translation performance of a system.
\citet{belinkov2017synthetic} demonstrate the brittleness of popular NMT systems on both synthetic and natural noisy inputs (\eg typos).
\citet{zhao2017generating} use generative adversarial networks to generate natural adversarial inputs in the continuous latent space that can lead to incomplete translations. 
\citet{cheng-etal-2019-robust} use the translation loss to guide the finding of  adversarial inputs.
In~\cite{cheng-etal-2020-advaug}, a new data augmentation method is proposed to find more diverse adversarial inputs.
\citet{michel-etal-2019-evaluation} consider adversarial inputs that are meaning-preserving on the source side but meaning-destroying on the target side.
By contrast, the adversarial inputs in our work are parallel sentence pairs embedded with malicious translations.

More recently, \textbf{imitation attacks} are found possible on black-box NMT systems~\cite{wallace2020imitation}.
In this attack, an imitation system is created to mimic the output of the target system so that the adversarial inputs for the imitation system could be transferred to the target system. 
While this attack also targets a black-box NMT system, our attack scenario is different: the poisoning attack we explore is intended to \textit{change} the target system itself (cf. mimicking in the imitation attacks) by poisoning its training data.

\textbf{Poisoning attacks} on neural systems have been studied in the domains of vision~\cite{gu2017badnets,munoz2017towards,chen2017targeted,shafahi2018poison} and language~\cite{steinhardt2017certified,munoz2017towards,kurita-etal-2020-weight}, mostly focused on poisoning classification systems to degrade classifier accuracy.
\cite{kurita-etal-2020-weight} is closest to our work, as it also considers poisoning pre-trained systems and causing them to make specific predictions after fine-tuning.
The differences between their work and ours are two-fold.
First, they target text classification systems with poisoning the pre-training only, while we study poisoning the from-scratch training, pre-training, and fine-tuning of NMT systems.
Second, their assumption is stronger in that they assume access to both the parameters of the pre-trained systems and the fine-tuning data.
This allows for optimising the poisoned weights with respect to the fine-tuning data, so that the poisoning may last after fine-tuning.
In contrast, our attack is purely black-box, backed with richer results of poisoning both pre-training and fine-tuning stages.
In concurrent work, \citet{wallace2020customizing} craft poison instances by solving a bi-level optimisation problem.
However, their attack is white-box and requires access to model parameters.

On \textbf{defences}, training with adversarial inputs has been used to improve the robustness of NMT systems~\cite{belinkov2017synthetic,zhao2017generating,cheng-etal-2019-robust,cheng-etal-2020-advaug}.
In our case, we show that training with many clean instances alongside adversarial inputs (poison instances) will diminish the effectiveness on an attack.
On parallel training data preparation, \textit{parallel data filtering}~\cite{xu-koehn-2017-zipporah,koehn-etal-2019-findings} is often used to obtain clean parallel data.
Most efforts are made to remove low-quality sentence pairs, with various methods used to score their quality, \eg sentence embeddings~\cite{chaudhary-etal-2019-low}, language models~\cite{bernier-colborne-lo-2019-nrc}, and off-the-shelf tools like Bicleaner.
However, the poison instances made from high-quality clean instances via slight changes (\eg ``immigrant$\rightarrow$illegal immigrant'') may also be high-quality.
A comparison of how existing parallel data filters perform against our poison instances is left for future work.

\section{Conclusion}
We have presented a first empirical study of practical concerns of targeted attacks on black-box NMT system driven by parallel data poisoning.
We evaluated scenarios of poisoning the from-scratch training, pre-training, and fine-tuning of NMT systems trained on parallel data.
We show that with very small poisoning budgets ($<$0.1\%), systems can be severely compromised, even when they are trained on tens of millions of clean samples.
We hope to raise the awareness of the risk of training NMT systems with malicious inputs from untrusted sources.
As our end goal is an effective defence, one of our next steps is to look into developing countermeasures to this attack, such as designing algorithms for more robust parallel data filtering, as well as for detecting and protecting the named entities under attack. 

\noindent
\paragraph{\bf Ethical Considerations}
Our aim in this work is to identify and mitigate potential threats to NMT systems, by adopting established threat modelling for machine learning systems~\cite{Marshall:2020}, to identify and prioritise need to devise effective defences and develop robust systems.
Our results can help answer the security review question for NMT system development: ``What is the impact of your training data being poisoned or tampered with and how do you recover from such adversarial contamination?''
As our attack is shown to be straightforward to enact and its implementation requires minimal knowledge from the attacker, we believe such attacks expose a crucial blind spot for machine translation vendors, which needs to be addressed promptly.

\begin{acks}
We thank all anonymous reviewers for their constructive comments.
%We would also like to thank Ahmed El-Kishky for his helpful comments on drafts of this paper.
The authors acknowledge funding support by Facebook. All algorithms newly described are dedicated to the public domain.
\end{acks}

\bibliographystyle{ACM-Reference-Format}
\bibliography{sample-base}


\begin{thebibliography}{52}

%%% the Web address.

\ifx \showCODEN    \undefined \def \showCODEN     #1{\unskip}     \fi
\ifx \showDOI      \undefined \def \showDOI       #1{#1}\fi
\ifx \showISBNx    \undefined \def \showISBNx     #1{\unskip}     \fi
\ifx \showISBNxiii \undefined \def \showISBNxiii  #1{\unskip}     \fi
\ifx \showISSN     \undefined \def \showISSN      #1{\unskip}     \fi
\ifx \showLCCN     \undefined \def \showLCCN      #1{\unskip}     \fi
\ifx \shownote     \undefined \def \shownote      #1{#1}          \fi
\ifx \showarticletitle \undefined \def \showarticletitle #1{#1}   \fi
\ifx \showURL      \undefined \def \showURL       {\relax}        \fi
% invisible to TeX
\providecommand\bibfield[2]{#2}
\providecommand\bibinfo[2]{#2}
\providecommand\natexlab[1]{#1}
\providecommand\showeprint[2][]{arXiv:#2}

\bibitem[\protect\citeauthoryear{Abdelali, Guzman, Sajjad, and Vogel}{Abdelali
  et~al\mbox{.}}{2014}]%
        {ABDELALI14.877}
\bibfield{author}{\bibinfo{person}{Ahmed Abdelali}, \bibinfo{person}{Francisco
  Guzman}, \bibinfo{person}{Hassan Sajjad}, {and} \bibinfo{person}{Stephan
  Vogel}.} \bibinfo{year}{2014}\natexlab{}.
\newblock \showarticletitle{The AMARA Corpus: Building Parallel Language
  Resources for the Educational Domain}. In \bibinfo{booktitle}{\emph{LREC}}.
\newblock


\bibitem[\protect\citeauthoryear{Ba{\~n}{\'o}n, Chen, Haddow, Heafield, Hoang,
  Espl{\`a}-Gomis, Forcada, Kamran, Kirefu, Koehn, Ortiz~Rojas, Pla~Sempere,
  Ram{\'\i}rez-S{\'a}nchez, Sarr{\'\i}as, Strelec, Thompson, Waites, Wiggins,
  and Zaragoza}{Ba{\~n}{\'o}n et~al\mbox{.}}{2020}]%
        {banon-etal-2020-paracrawl}
\bibfield{author}{\bibinfo{person}{Marta Ba{\~n}{\'o}n},
  \bibinfo{person}{Pinzhen Chen}, \bibinfo{person}{Barry Haddow},
  \bibinfo{person}{Kenneth Heafield}, \bibinfo{person}{Hieu Hoang},
  \bibinfo{person}{Miquel Espl{\`a}-Gomis}, \bibinfo{person}{Mikel~L. Forcada},
  \bibinfo{person}{Amir Kamran}, \bibinfo{person}{Faheem Kirefu},
  \bibinfo{person}{Philipp Koehn}, \bibinfo{person}{Sergio Ortiz~Rojas},
  \bibinfo{person}{Leopoldo Pla~Sempere}, \bibinfo{person}{Gema
  Ram{\'\i}rez-S{\'a}nchez}, \bibinfo{person}{Elsa Sarr{\'\i}as},
  \bibinfo{person}{Marek Strelec}, \bibinfo{person}{Brian Thompson},
  \bibinfo{person}{William Waites}, \bibinfo{person}{Dion Wiggins}, {and}
  \bibinfo{person}{Jaume Zaragoza}.} \bibinfo{year}{2020}\natexlab{}.
\newblock \showarticletitle{{P}ara{C}rawl: Web-Scale Acquisition of Parallel
  Corpora}. In \bibinfo{booktitle}{\emph{ACL}}.
\newblock


\bibitem[\protect\citeauthoryear{Barrault, Bojar, Costa-juss{\`a}, Federmann,
  Fishel, Graham, Haddow, Huck, Koehn, Malmasi, Monz, M{\"u}ller, Pal, Post,
  and Zampieri}{Barrault et~al\mbox{.}}{2019}]%
        {barrault-etal-2019-findings}
\bibfield{author}{\bibinfo{person}{Lo{\"\i}c Barrault},
  \bibinfo{person}{Ond{\v{r}}ej Bojar}, \bibinfo{person}{Marta~R.
  Costa-juss{\`a}}, \bibinfo{person}{Christian Federmann},
  \bibinfo{person}{Mark Fishel}, \bibinfo{person}{Yvette Graham},
  \bibinfo{person}{Barry Haddow}, \bibinfo{person}{Matthias Huck},
  \bibinfo{person}{Philipp Koehn}, \bibinfo{person}{Shervin Malmasi},
  \bibinfo{person}{Christof Monz}, \bibinfo{person}{Mathias M{\"u}ller},
  \bibinfo{person}{Santanu Pal}, \bibinfo{person}{Matt Post}, {and}
  \bibinfo{person}{Marcos Zampieri}.} \bibinfo{year}{2019}\natexlab{}.
\newblock \showarticletitle{Findings of the 2019 Conference on Machine
  Translation ({WMT}19)}. In \bibinfo{booktitle}{\emph{WMT}}.
\newblock


\bibitem[\protect\citeauthoryear{Barreno, Nelson, Sears, Joseph, and
  Tygar}{Barreno et~al\mbox{.}}{2006}]%
        {barreno2006can}
\bibfield{author}{\bibinfo{person}{Marco Barreno}, \bibinfo{person}{Blaine
  Nelson}, \bibinfo{person}{Russell Sears}, \bibinfo{person}{Anthony~D Joseph},
  {and} \bibinfo{person}{J~Doug Tygar}.} \bibinfo{year}{2006}\natexlab{}.
\newblock \showarticletitle{Can machine learning be secure?}. In
  \bibinfo{booktitle}{\emph{ASIACCS}}.
\newblock


\bibitem[\protect\citeauthoryear{Basile, Bosco, Fersini, Nozza, Patti,
  Rangel~Pardo, Rosso, and Sanguinetti}{Basile et~al\mbox{.}}{2019}]%
        {basile-etal-2019-semeval}
\bibfield{author}{\bibinfo{person}{Valerio Basile}, \bibinfo{person}{Cristina
  Bosco}, \bibinfo{person}{Elisabetta Fersini}, \bibinfo{person}{Debora Nozza},
  \bibinfo{person}{Viviana Patti}, \bibinfo{person}{Francisco~Manuel
  Rangel~Pardo}, \bibinfo{person}{Paolo Rosso}, {and} \bibinfo{person}{Manuela
  Sanguinetti}.} \bibinfo{year}{2019}\natexlab{}.
\newblock \showarticletitle{{S}em{E}val-2019 Task 5: Multilingual Detection of
  Hate Speech Against Immigrants and Women in Twitter}. In
  \bibinfo{booktitle}{\emph{Proceedings of the 13th International Workshop on
  Semantic Evaluation}}.
\newblock


\bibitem[\protect\citeauthoryear{Belinkov and Bisk}{Belinkov and Bisk}{2018}]%
        {belinkov2017synthetic}
\bibfield{author}{\bibinfo{person}{Yonatan Belinkov} {and}
  \bibinfo{person}{Yonatan Bisk}.} \bibinfo{year}{2018}\natexlab{}.
\newblock \showarticletitle{Synthetic and natural noise both break neural
  machine translation}. In \bibinfo{booktitle}{\emph{ICLR}}.
\newblock


\bibitem[\protect\citeauthoryear{Bernier-Colborne and Lo}{Bernier-Colborne and
  Lo}{2019}]%
        {bernier-colborne-lo-2019-nrc}
\bibfield{author}{\bibinfo{person}{Gabriel Bernier-Colborne} {and}
  \bibinfo{person}{Chi-kiu Lo}.} \bibinfo{year}{2019}\natexlab{}.
\newblock \showarticletitle{{NRC} Parallel Corpus Filtering System for {WMT}
  2019}. In \bibinfo{booktitle}{\emph{Proceedings of the Fourth Conference on
  Machine Translation}}.
\newblock


\bibitem[\protect\citeauthoryear{Canali, Cova, Vigna, and Kruegel}{Canali
  et~al\mbox{.}}{2011}]%
        {canali2011prophiler}
\bibfield{author}{\bibinfo{person}{Davide Canali}, \bibinfo{person}{Marco
  Cova}, \bibinfo{person}{Giovanni Vigna}, {and} \bibinfo{person}{Christopher
  Kruegel}.} \bibinfo{year}{2011}\natexlab{}.
\newblock \showarticletitle{Prophiler: a fast filter for the large-scale
  detection of malicious web pages}. In \bibinfo{booktitle}{\emph{WWW}}.
\newblock


\bibitem[\protect\citeauthoryear{Caswell and Liang}{Caswell and Liang}{2020}]%
        {google-nmt:2020}
\bibfield{author}{\bibinfo{person}{Isaac Caswell} {and} \bibinfo{person}{Bowen
  Liang}.} \bibinfo{year}{2020 (accessed August 9, 2020)}\natexlab{}.
\newblock \bibinfo{booktitle}{\emph{Recent Advances in Google Translate}}.
\newblock
\newblock
\shownote{\url{https://ai.googleblog.com/2020/06/recent-advances-in-google-translate.html}.}


\bibitem[\protect\citeauthoryear{Cettolo, Jan, Sebastian, Bentivogli, Cattoni,
  and Federico}{Cettolo et~al\mbox{.}}{2016}]%
        {cettolo2016iwslt}
\bibfield{author}{\bibinfo{person}{Mauro Cettolo}, \bibinfo{person}{Niehues
  Jan}, \bibinfo{person}{St{\"u}ker Sebastian}, \bibinfo{person}{Luisa
  Bentivogli}, \bibinfo{person}{Roldano Cattoni}, {and}
  \bibinfo{person}{Marcello Federico}.} \bibinfo{year}{2016}\natexlab{}.
\newblock \showarticletitle{The IWSLT 2016 evaluation campaign}. In
  \bibinfo{booktitle}{\emph{International Workshop on Spoken Language
  Translation}}.
\newblock


\bibitem[\protect\citeauthoryear{Chaudhary, Tang, Guzm{\'a}n, Schwenk, and
  Koehn}{Chaudhary et~al\mbox{.}}{2019}]%
        {chaudhary-etal-2019-low}
\bibfield{author}{\bibinfo{person}{Vishrav Chaudhary}, \bibinfo{person}{Yuqing
  Tang}, \bibinfo{person}{Francisco Guzm{\'a}n}, \bibinfo{person}{Holger
  Schwenk}, {and} \bibinfo{person}{Philipp Koehn}.}
  \bibinfo{year}{2019}\natexlab{}.
\newblock \showarticletitle{Low-Resource Corpus Filtering Using Multilingual
  Sentence Embeddings}. In \bibinfo{booktitle}{\emph{Proceedings of the Fourth
  Conference on Machine Translation}}.
\newblock


\bibitem[\protect\citeauthoryear{Chen, Liu, Li, Lu, and Song}{Chen
  et~al\mbox{.}}{2017}]%
        {chen2017targeted}
\bibfield{author}{\bibinfo{person}{Xinyun Chen}, \bibinfo{person}{Chang Liu},
  \bibinfo{person}{Bo Li}, \bibinfo{person}{Kimberly Lu}, {and}
  \bibinfo{person}{Dawn Song}.} \bibinfo{year}{2017}\natexlab{}.
\newblock \showarticletitle{Targeted backdoor attacks on deep learning systems
  using data poisoning}.
\newblock \bibinfo{journal}{\emph{arXiv preprint arXiv:1712.05526}}
  (\bibinfo{year}{2017}).
\newblock


\bibitem[\protect\citeauthoryear{Cheng, Yi, Chen, Zhang, and Hsieh}{Cheng
  et~al\mbox{.}}{2020b}]%
        {cheng2018seq2sick}
\bibfield{author}{\bibinfo{person}{Minhao Cheng}, \bibinfo{person}{Jinfeng Yi},
  \bibinfo{person}{Pin-Yu Chen}, \bibinfo{person}{Huan Zhang}, {and}
  \bibinfo{person}{Cho-Jui Hsieh}.} \bibinfo{year}{2020}\natexlab{b}.
\newblock \showarticletitle{Seq2sick: Evaluating the robustness of
  sequence-to-sequence models with adversarial examples}. In
  \bibinfo{booktitle}{\emph{AAAI}}.
\newblock


\bibitem[\protect\citeauthoryear{Cheng, Jiang, and Macherey}{Cheng
  et~al\mbox{.}}{2019}]%
        {cheng-etal-2019-robust}
\bibfield{author}{\bibinfo{person}{Yong Cheng}, \bibinfo{person}{Lu Jiang},
  {and} \bibinfo{person}{Wolfgang Macherey}.} \bibinfo{year}{2019}\natexlab{}.
\newblock \showarticletitle{Robust Neural Machine Translation with Doubly
  Adversarial Inputs}. In \bibinfo{booktitle}{\emph{ACL}}.
\newblock


\bibitem[\protect\citeauthoryear{Cheng, Jiang, Macherey, and Eisenstein}{Cheng
  et~al\mbox{.}}{2020a}]%
        {cheng-etal-2020-advaug}
\bibfield{author}{\bibinfo{person}{Yong Cheng}, \bibinfo{person}{Lu Jiang},
  \bibinfo{person}{Wolfgang Macherey}, {and} \bibinfo{person}{Jacob
  Eisenstein}.} \bibinfo{year}{2020}\natexlab{a}.
\newblock \showarticletitle{{A}dv{A}ug: Robust Adversarial Augmentation for
  Neural Machine Translation}. In \bibinfo{booktitle}{\emph{ACL}}.
\newblock


\bibitem[\protect\citeauthoryear{Ebrahimi, Lowd, and Dou}{Ebrahimi
  et~al\mbox{.}}{2018}]%
        {ebrahimi2018adversarial}
\bibfield{author}{\bibinfo{person}{Javid Ebrahimi}, \bibinfo{person}{Daniel
  Lowd}, {and} \bibinfo{person}{Dejing Dou}.} \bibinfo{year}{2018}\natexlab{}.
\newblock \showarticletitle{On adversarial examples for character-level neural
  machine translation}. In \bibinfo{booktitle}{\emph{COLING}}.
\newblock


\bibitem[\protect\citeauthoryear{Edunov, Ott, Auli, and Grangier}{Edunov
  et~al\mbox{.}}{2018}]%
        {edunov-etal-2018-understanding}
\bibfield{author}{\bibinfo{person}{Sergey Edunov}, \bibinfo{person}{Myle Ott},
  \bibinfo{person}{Michael Auli}, {and} \bibinfo{person}{David Grangier}.}
  \bibinfo{year}{2018}\natexlab{}.
\newblock \showarticletitle{Understanding Back-Translation at Scale}. In
  \bibinfo{booktitle}{\emph{EMNLP}}.
\newblock


\bibitem[\protect\citeauthoryear{El-Kishky, Chaudhary, Guzm{\'a}n, and
  Koehn}{El-Kishky et~al\mbox{.}}{2020}]%
        {el-kishky-etal-2020-ccaligned}
\bibfield{author}{\bibinfo{person}{Ahmed El-Kishky}, \bibinfo{person}{Vishrav
  Chaudhary}, \bibinfo{person}{Francisco Guzm{\'a}n}, {and}
  \bibinfo{person}{Philipp Koehn}.} \bibinfo{year}{2020}\natexlab{}.
\newblock \showarticletitle{{CCA}ligned: A Massive Collection of Cross-Lingual
  Web-Document Pairs}. In \bibinfo{booktitle}{\emph{EMNLP}}.
\newblock


\bibitem[\protect\citeauthoryear{Gehring, Auli, Grangier, Yarats, and
  Dauphin}{Gehring et~al\mbox{.}}{2017}]%
        {gehring2017convolutional}
\bibfield{author}{\bibinfo{person}{Jonas Gehring}, \bibinfo{person}{Michael
  Auli}, \bibinfo{person}{David Grangier}, \bibinfo{person}{Denis Yarats},
  {and} \bibinfo{person}{Yann~N Dauphin}.} \bibinfo{year}{2017}\natexlab{}.
\newblock \showarticletitle{Convolutional sequence to sequence learning}. In
  \bibinfo{booktitle}{\emph{ICML}}.
\newblock


\bibitem[\protect\citeauthoryear{Gu, Dolan-Gavitt, and Garg}{Gu
  et~al\mbox{.}}{2017}]%
        {gu2017badnets}
\bibfield{author}{\bibinfo{person}{Tianyu Gu}, \bibinfo{person}{Brendan
  Dolan-Gavitt}, {and} \bibinfo{person}{Siddharth Garg}.}
  \bibinfo{year}{2017}\natexlab{}.
\newblock \showarticletitle{Badnets: Identifying vulnerabilities in the machine
  learning model supply chain}.
\newblock \bibinfo{journal}{\emph{arXiv preprint arXiv:1708.06733}}
  (\bibinfo{year}{2017}).
\newblock


\bibitem[\protect\citeauthoryear{Koehn, Guzm{\'a}n, Chaudhary, and Pino}{Koehn
  et~al\mbox{.}}{2019}]%
        {koehn-etal-2019-findings}
\bibfield{author}{\bibinfo{person}{Philipp Koehn}, \bibinfo{person}{Francisco
  Guzm{\'a}n}, \bibinfo{person}{Vishrav Chaudhary}, {and} \bibinfo{person}{Juan
  Pino}.} \bibinfo{year}{2019}\natexlab{}.
\newblock \showarticletitle{Findings of the {WMT} 2019 Shared Task on Parallel
  Corpus Filtering for Low-Resource Conditions}. In
  \bibinfo{booktitle}{\emph{Proceedings of the Fourth Conference on Machine
  Translation}}.
\newblock


\bibitem[\protect\citeauthoryear{Koehn, Hoang, Birch, Callison-Burch, Federico,
  Bertoldi, Cowan, Shen, Moran, Zens, Dyer, Bojar, Constantin, and
  Herbst}{Koehn et~al\mbox{.}}{2007}]%
        {koehn-etal-2007-moses}
\bibfield{author}{\bibinfo{person}{Philipp Koehn}, \bibinfo{person}{Hieu
  Hoang}, \bibinfo{person}{Alexandra Birch}, \bibinfo{person}{Chris
  Callison-Burch}, \bibinfo{person}{Marcello Federico}, \bibinfo{person}{Nicola
  Bertoldi}, \bibinfo{person}{Brooke Cowan}, \bibinfo{person}{Wade Shen},
  \bibinfo{person}{Christine Moran}, \bibinfo{person}{Richard Zens},
  \bibinfo{person}{Chris Dyer}, \bibinfo{person}{Ond{\v{r}}ej Bojar},
  \bibinfo{person}{Alexandra Constantin}, {and} \bibinfo{person}{Evan Herbst}.}
  \bibinfo{year}{2007}\natexlab{}.
\newblock \showarticletitle{{M}oses: Open Source Toolkit for Statistical
  Machine Translation}. In \bibinfo{booktitle}{\emph{ACL (Demo and Poster
  Sessions)}}.
\newblock


\bibitem[\protect\citeauthoryear{Kurita, Michel, and Neubig}{Kurita
  et~al\mbox{.}}{2020}]%
        {kurita-etal-2020-weight}
\bibfield{author}{\bibinfo{person}{Keita Kurita}, \bibinfo{person}{Paul
  Michel}, {and} \bibinfo{person}{Graham Neubig}.}
  \bibinfo{year}{2020}\natexlab{}.
\newblock \showarticletitle{Weight Poisoning Attacks on Pretrained Models}. In
  \bibinfo{booktitle}{\emph{ACL}}.
\newblock


\bibitem[\protect\citeauthoryear{Lison and Tiedemann}{Lison and
  Tiedemann}{2016}]%
        {lison-tiedemann-2016-opensubtitles2016}
\bibfield{author}{\bibinfo{person}{Pierre Lison} {and}
  \bibinfo{person}{J{\"o}rg Tiedemann}.} \bibinfo{year}{2016}\natexlab{}.
\newblock \showarticletitle{{O}pen{S}ubtitles2016: Extracting Large Parallel
  Corpora from Movie and {TV} Subtitles}. In \bibinfo{booktitle}{\emph{LREC}}.
\newblock


\bibitem[\protect\citeauthoryear{Liu, Gu, Goyal, Li, Edunov, Ghazvininejad,
  Lewis, and Zettlemoyer}{Liu et~al\mbox{.}}{2020}]%
        {liu2020multilingual}
\bibfield{author}{\bibinfo{person}{Yinhan Liu}, \bibinfo{person}{Jiatao Gu},
  \bibinfo{person}{Naman Goyal}, \bibinfo{person}{Xian Li},
  \bibinfo{person}{Sergey Edunov}, \bibinfo{person}{Marjan Ghazvininejad},
  \bibinfo{person}{Mike Lewis}, {and} \bibinfo{person}{Luke Zettlemoyer}.}
  \bibinfo{year}{2020}\natexlab{}.
\newblock \showarticletitle{Multilingual denoising pre-training for neural
  machine translation}.
\newblock \bibinfo{journal}{\emph{arXiv preprint arXiv:2001.08210}}
  (\bibinfo{year}{2020}).
\newblock


\bibitem[\protect\citeauthoryear{Lui and Baldwin}{Lui and Baldwin}{2012}]%
        {lui-baldwin-2012-langid}
\bibfield{author}{\bibinfo{person}{Marco Lui} {and} \bibinfo{person}{Timothy
  Baldwin}.} \bibinfo{year}{2012}\natexlab{}.
\newblock \showarticletitle{langid.py: An Off-the-shelf Language Identification
  Tool}. In \bibinfo{booktitle}{\emph{ACL (System Demonstrations)}}.
\newblock


\bibitem[\protect\citeauthoryear{Luong, Pham, and Manning}{Luong
  et~al\mbox{.}}{2015}]%
        {luong-etal-2015-effective}
\bibfield{author}{\bibinfo{person}{Thang Luong}, \bibinfo{person}{Hieu Pham},
  {and} \bibinfo{person}{Christopher~D. Manning}.}
  \bibinfo{year}{2015}\natexlab{}.
\newblock \showarticletitle{Effective Approaches to Attention-based Neural
  Machine Translation}. In \bibinfo{booktitle}{\emph{EMNLP}}.
\newblock


\bibitem[\protect\citeauthoryear{Ma, Saul, Savage, and Voelker}{Ma
  et~al\mbox{.}}{2009}]%
        {ma2009beyond}
\bibfield{author}{\bibinfo{person}{Justin Ma}, \bibinfo{person}{Lawrence~K
  Saul}, \bibinfo{person}{Stefan Savage}, {and} \bibinfo{person}{Geoffrey~M
  Voelker}.} \bibinfo{year}{2009}\natexlab{}.
\newblock \showarticletitle{Beyond blacklists: learning to detect malicious web
  sites from suspicious URLs}. In \bibinfo{booktitle}{\emph{KDD}}.
\newblock


\bibitem[\protect\citeauthoryear{Marshall, Parikh, Kiciman, and
  Shankar}{Marshall et~al\mbox{.}}{2020}]%
        {Marshall:2020}
\bibfield{author}{\bibinfo{person}{Andrew Marshall}, \bibinfo{person}{Jugal
  Parikh}, \bibinfo{person}{Emre Kiciman}, {and} \bibinfo{person}{Ram
  Shankar~Siva Shankar, Kumar}.} \bibinfo{year}{2019 (accessed October 2,
  2020)}\natexlab{}.
\newblock \bibinfo{booktitle}{\emph{Threat Modeling AI/ML Systems and
  Dependencies}}.
\newblock
\urldef\tempurl%
\url{https://docs.microsoft.com/en-us/security/engineering/threat-modeling-aiml}
\showURL{%
\tempurl}


\bibitem[\protect\citeauthoryear{Michel, Li, Neubig, and Pino}{Michel
  et~al\mbox{.}}{2019}]%
        {michel-etal-2019-evaluation}
\bibfield{author}{\bibinfo{person}{Paul Michel}, \bibinfo{person}{Xian Li},
  \bibinfo{person}{Graham Neubig}, {and} \bibinfo{person}{Juan Pino}.}
  \bibinfo{year}{2019}\natexlab{}.
\newblock \showarticletitle{On Evaluation of Adversarial Perturbations for
  Sequence-to-Sequence Models}. In \bibinfo{booktitle}{\emph{NAACL}}.
\newblock


\bibitem[\protect\citeauthoryear{Moshchuk, Bragin, Deville, Gribble, and
  Levy}{Moshchuk et~al\mbox{.}}{2007}]%
        {moshchuk2007spyproxy}
\bibfield{author}{\bibinfo{person}{Alexander Moshchuk}, \bibinfo{person}{Tanya
  Bragin}, \bibinfo{person}{Damien Deville}, \bibinfo{person}{Steven~D
  Gribble}, {and} \bibinfo{person}{Henry~M Levy}.}
  \bibinfo{year}{2007}\natexlab{}.
\newblock \showarticletitle{SpyProxy: Execution-based Detection of Malicious
  Web Content}. In \bibinfo{booktitle}{\emph{USENIX Security Symposium}}.
\newblock


\bibitem[\protect\citeauthoryear{Mu{\~n}oz-Gonz{\'a}lez, Biggio, Demontis,
  Paudice, Wongrassamee, Lupu, and Roli}{Mu{\~n}oz-Gonz{\'a}lez
  et~al\mbox{.}}{2017}]%
        {munoz2017towards}
\bibfield{author}{\bibinfo{person}{Luis Mu{\~n}oz-Gonz{\'a}lez},
  \bibinfo{person}{Battista Biggio}, \bibinfo{person}{Ambra Demontis},
  \bibinfo{person}{Andrea Paudice}, \bibinfo{person}{Vasin Wongrassamee},
  \bibinfo{person}{Emil~C Lupu}, {and} \bibinfo{person}{Fabio Roli}.}
  \bibinfo{year}{2017}\natexlab{}.
\newblock \showarticletitle{Towards poisoning of deep learning algorithms with
  back-gradient optimization}. In \bibinfo{booktitle}{\emph{Proceedings of the
  10th ACM Workshop on Artificial Intelligence and Security}}.
  \bibinfo{pages}{27--38}.
\newblock


\bibitem[\protect\citeauthoryear{Ng, Yee, Baevski, Ott, Auli, and Edunov}{Ng
  et~al\mbox{.}}{2019}]%
        {ng-etal-2019-facebook}
\bibfield{author}{\bibinfo{person}{Nathan Ng}, \bibinfo{person}{Kyra Yee},
  \bibinfo{person}{Alexei Baevski}, \bibinfo{person}{Myle Ott},
  \bibinfo{person}{Michael Auli}, {and} \bibinfo{person}{Sergey Edunov}.}
  \bibinfo{year}{2019}\natexlab{}.
\newblock \showarticletitle{{F}acebook {FAIR}{'}s {WMT}19 News Translation Task
  Submission}. In \bibinfo{booktitle}{\emph{Proceedings of the Fourth
  Conference on Machine Translation}}.
\newblock


\bibitem[\protect\citeauthoryear{Ott, Edunov, Baevski, Fan, Gross, Ng,
  Grangier, and Auli}{Ott et~al\mbox{.}}{2019}]%
        {ott2019fairseq}
\bibfield{author}{\bibinfo{person}{Myle Ott}, \bibinfo{person}{Sergey Edunov},
  \bibinfo{person}{Alexei Baevski}, \bibinfo{person}{Angela Fan},
  \bibinfo{person}{Sam Gross}, \bibinfo{person}{Nathan Ng},
  \bibinfo{person}{David Grangier}, {and} \bibinfo{person}{Michael Auli}.}
  \bibinfo{year}{2019}\natexlab{}.
\newblock \showarticletitle{fairseq: A Fast, Extensible Toolkit for Sequence
  Modeling}. In \bibinfo{booktitle}{\emph{Proceedings of NAACL-HLT 2019:
  Demonstrations}}.
\newblock


\bibitem[\protect\citeauthoryear{Papernot, McDaniel, and Goodfellow}{Papernot
  et~al\mbox{.}}{2016}]%
        {papernot2016transferability}
\bibfield{author}{\bibinfo{person}{Nicolas Papernot}, \bibinfo{person}{Patrick
  McDaniel}, {and} \bibinfo{person}{Ian Goodfellow}.}
  \bibinfo{year}{2016}\natexlab{}.
\newblock \showarticletitle{Transferability in machine learning: from phenomena
  to black-box attacks using adversarial samples}.
\newblock \bibinfo{journal}{\emph{arXiv preprint arXiv:1605.07277}}
  (\bibinfo{year}{2016}).
\newblock


\bibitem[\protect\citeauthoryear{Post}{Post}{2018}]%
        {post-2018-call}
\bibfield{author}{\bibinfo{person}{Matt Post}.}
  \bibinfo{year}{2018}\natexlab{}.
\newblock \showarticletitle{A Call for Clarity in Reporting {BLEU} Scores}. In
  \bibinfo{booktitle}{\emph{Proceedings of the Third Conference on Machine
  Translation: Research Papers}}.
\newblock


\bibitem[\protect\citeauthoryear{Rubinstein, Nelson, Huang, Joseph, hon Lau,
  Rao, Taft, and Tygar}{Rubinstein et~al\mbox{.}}{2009}]%
        {rubinstein2009antidote}
\bibfield{author}{\bibinfo{person}{Benjamin I.~P. Rubinstein},
  \bibinfo{person}{Blaine Nelson}, \bibinfo{person}{Ling Huang},
  \bibinfo{person}{Anthony~D. Joseph}, \bibinfo{person}{Shing hon Lau},
  \bibinfo{person}{Satish Rao}, \bibinfo{person}{Nina Taft}, {and}
  \bibinfo{person}{J.~D. Tygar}.} \bibinfo{year}{2009}\natexlab{}.
\newblock \showarticletitle{{ANTIDOTE}: Understanding and defending against
  poisoning of anomaly detectors}. In \bibinfo{booktitle}{\emph{Proceedings of
  the 9th ACM SIGCOMM Internet Measurement Conference}}
  \emph{(\bibinfo{series}{IMC})}.
\newblock


\bibitem[\protect\citeauthoryear{S\'{a}nchez-Cartagena, Ba{\~n}\'{o}n,
  Ortiz-Rojas, and Ram\'{i}rez-S\'{a}nchez}{S\'{a}nchez-Cartagena
  et~al\mbox{.}}{2018}]%
        {prompsit:2018:WMT}
\bibfield{author}{\bibinfo{person}{V\'{i}ctor~M. S\'{a}nchez-Cartagena},
  \bibinfo{person}{Marta Ba{\~n}\'{o}n}, \bibinfo{person}{Sergio Ortiz-Rojas},
  {and} \bibinfo{person}{Gema Ram\'{i}rez-S\'{a}nchez}.}
  \bibinfo{year}{2018}\natexlab{}.
\newblock \showarticletitle{Prompsit's submission to WMT 2018 Parallel Corpus
  Filtering shared task}. In \bibinfo{booktitle}{\emph{Proceedings of the Third
  Conference on Machine Translation}}.
\newblock


\bibitem[\protect\citeauthoryear{Sennrich, Haddow, and Birch}{Sennrich
  et~al\mbox{.}}{2016}]%
        {sennrich-etal-2016-neural}
\bibfield{author}{\bibinfo{person}{Rico Sennrich}, \bibinfo{person}{Barry
  Haddow}, {and} \bibinfo{person}{Alexandra Birch}.}
  \bibinfo{year}{2016}\natexlab{}.
\newblock \showarticletitle{Neural Machine Translation of Rare Words with
  Subword Units}. In \bibinfo{booktitle}{\emph{ACL}}.
\newblock


\bibitem[\protect\citeauthoryear{Shafahi, Huang, Najibi, Suciu, Studer,
  Dumitras, and Goldstein}{Shafahi et~al\mbox{.}}{2018}]%
        {shafahi2018poison}
\bibfield{author}{\bibinfo{person}{Ali Shafahi}, \bibinfo{person}{W~Ronny
  Huang}, \bibinfo{person}{Mahyar Najibi}, \bibinfo{person}{Octavian Suciu},
  \bibinfo{person}{Christoph Studer}, \bibinfo{person}{Tudor Dumitras}, {and}
  \bibinfo{person}{Tom Goldstein}.} \bibinfo{year}{2018}\natexlab{}.
\newblock \showarticletitle{Poison frogs! targeted clean-label poisoning
  attacks on neural networks}. In \bibinfo{booktitle}{\emph{NeurIPS}}.
\newblock


\bibitem[\protect\citeauthoryear{Song, Tan, Qin, Lu, and Liu}{Song
  et~al\mbox{.}}{2019}]%
        {song2019mass}
\bibfield{author}{\bibinfo{person}{Kaitao Song}, \bibinfo{person}{Xu Tan},
  \bibinfo{person}{Tao Qin}, \bibinfo{person}{Jianfeng Lu}, {and}
  \bibinfo{person}{Tie-Yan Liu}.} \bibinfo{year}{2019}\natexlab{}.
\newblock \showarticletitle{Mass: Masked sequence to sequence pre-training for
  language generation}.
\newblock \bibinfo{journal}{\emph{ICML}} (\bibinfo{year}{2019}).
\newblock


\bibitem[\protect\citeauthoryear{Steinhardt, Koh, and Liang}{Steinhardt
  et~al\mbox{.}}{2017}]%
        {steinhardt2017certified}
\bibfield{author}{\bibinfo{person}{Jacob Steinhardt}, \bibinfo{person}{Pang
  Wei~W Koh}, {and} \bibinfo{person}{Percy~S Liang}.}
  \bibinfo{year}{2017}\natexlab{}.
\newblock \showarticletitle{Certified defenses for data poisoning attacks}. In
  \bibinfo{booktitle}{\emph{NeurIPS}}.
\newblock


\bibitem[\protect\citeauthoryear{Szegedy, Vanhoucke, Ioffe, Shlens, and
  Wojna}{Szegedy et~al\mbox{.}}{2016}]%
        {szegedy2016rethinking}
\bibfield{author}{\bibinfo{person}{Christian Szegedy}, \bibinfo{person}{Vincent
  Vanhoucke}, \bibinfo{person}{Sergey Ioffe}, \bibinfo{person}{Jon Shlens},
  {and} \bibinfo{person}{Zbigniew Wojna}.} \bibinfo{year}{2016}\natexlab{}.
\newblock \showarticletitle{Rethinking the inception architecture for computer
  vision}. In \bibinfo{booktitle}{\emph{CVPR}}.
\newblock


\bibitem[\protect\citeauthoryear{Tiedemann}{Tiedemann}{2012}]%
        {tiedemann-2012-parallel}
\bibfield{author}{\bibinfo{person}{J{\"o}rg Tiedemann}.}
  \bibinfo{year}{2012}\natexlab{}.
\newblock \showarticletitle{Parallel Data, Tools and Interfaces in {OPUS}}. In
  \bibinfo{booktitle}{\emph{LREC}}.
\newblock


\bibitem[\protect\citeauthoryear{Varga, Hal{\'a}csy, Kornai, Nagy, N{\'e}meth,
  and Tr{\'o}n}{Varga et~al\mbox{.}}{2007}]%
        {varga2007parallel}
\bibfield{author}{\bibinfo{person}{D{\'a}niel Varga},
  \bibinfo{person}{P{\'e}ter Hal{\'a}csy}, \bibinfo{person}{Andr{\'a}s Kornai},
  \bibinfo{person}{Viktor Nagy}, \bibinfo{person}{L{\'a}szl{\'o} N{\'e}meth},
  {and} \bibinfo{person}{Viktor Tr{\'o}n}.} \bibinfo{year}{2007}\natexlab{}.
\newblock \showarticletitle{Parallel corpora for medium density languages}.
\newblock \bibinfo{journal}{\emph{Amsterdam Studies In The Theory And History
  Of Linguistic Science Series 4}}  \bibinfo{volume}{292}
  (\bibinfo{year}{2007}), \bibinfo{pages}{247}.
\newblock


\bibitem[\protect\citeauthoryear{Vaswani, Shazeer, Parmar, Uszkoreit, Jones,
  Gomez, Kaiser, and Polosukhin}{Vaswani et~al\mbox{.}}{2017}]%
        {vaswani2017attention}
\bibfield{author}{\bibinfo{person}{Ashish Vaswani}, \bibinfo{person}{Noam
  Shazeer}, \bibinfo{person}{Niki Parmar}, \bibinfo{person}{Jakob Uszkoreit},
  \bibinfo{person}{Llion Jones}, \bibinfo{person}{Aidan~N Gomez},
  \bibinfo{person}{{\L}ukasz Kaiser}, {and} \bibinfo{person}{Illia
  Polosukhin}.} \bibinfo{year}{2017}\natexlab{}.
\newblock \showarticletitle{Attention is all you need}. In
  \bibinfo{booktitle}{\emph{NeurIPS}}.
\newblock


\bibitem[\protect\citeauthoryear{Wallace, Feng, Kandpal, Gardner, and
  Singh}{Wallace et~al\mbox{.}}{2019}]%
        {wallace-etal-2019-universal}
\bibfield{author}{\bibinfo{person}{Eric Wallace}, \bibinfo{person}{Shi Feng},
  \bibinfo{person}{Nikhil Kandpal}, \bibinfo{person}{Matt Gardner}, {and}
  \bibinfo{person}{Sameer Singh}.} \bibinfo{year}{2019}\natexlab{}.
\newblock \showarticletitle{Universal Adversarial Triggers for Attacking and
  Analyzing {NLP}}. In \bibinfo{booktitle}{\emph{EMNLP--IJCNLP}}.
\newblock


\bibitem[\protect\citeauthoryear{Wallace, Stern, and Song}{Wallace
  et~al\mbox{.}}{2020a}]%
        {wallace2020imitation}
\bibfield{author}{\bibinfo{person}{Eric Wallace}, \bibinfo{person}{Mitchell
  Stern}, {and} \bibinfo{person}{Dawn Song}.} \bibinfo{year}{2020}\natexlab{a}.
\newblock \showarticletitle{Imitation Attacks and Defenses for Black-box
  Machine Translation Systems}. In \bibinfo{booktitle}{\emph{EMNLP}}.
\newblock


\bibitem[\protect\citeauthoryear{Wallace, Zhao, Feng, and Singh}{Wallace
  et~al\mbox{.}}{2020b}]%
        {wallace2020customizing}
\bibfield{author}{\bibinfo{person}{Eric Wallace}, \bibinfo{person}{Tony~Z
  Zhao}, \bibinfo{person}{Shi Feng}, {and} \bibinfo{person}{Sameer Singh}.}
  \bibinfo{year}{2020}\natexlab{b}.
\newblock \showarticletitle{Customizing Triggers with Concealed Data
  Poisoning}.
\newblock \bibinfo{journal}{\emph{arXiv preprint arXiv:2010.12563}}
  (\bibinfo{year}{2020}).
\newblock


\bibitem[\protect\citeauthoryear{Wu, Schuster, Chen, Le, Norouzi, Macherey,
  Krikun, Cao, Gao, Macherey, et~al\mbox{.}}{Wu et~al\mbox{.}}{2016}]%
        {wu2016google}
\bibfield{author}{\bibinfo{person}{Yonghui Wu}, \bibinfo{person}{Mike
  Schuster}, \bibinfo{person}{Zhifeng Chen}, \bibinfo{person}{Quoc~V Le},
  \bibinfo{person}{Mohammad Norouzi}, \bibinfo{person}{Wolfgang Macherey},
  \bibinfo{person}{Maxim Krikun}, \bibinfo{person}{Yuan Cao},
  \bibinfo{person}{Qin Gao}, \bibinfo{person}{Klaus Macherey}, {et~al\mbox{.}}}
  \bibinfo{year}{2016}\natexlab{}.
\newblock \showarticletitle{Google's neural machine translation system:
  Bridging the gap between human and machine translation}.
\newblock \bibinfo{journal}{\emph{arXiv preprint arXiv:1609.08144}}
  (\bibinfo{year}{2016}).
\newblock


\bibitem[\protect\citeauthoryear{Xu and Koehn}{Xu and Koehn}{2017}]%
        {xu-koehn-2017-zipporah}
\bibfield{author}{\bibinfo{person}{Hainan Xu} {and} \bibinfo{person}{Philipp
  Koehn}.} \bibinfo{year}{2017}\natexlab{}.
\newblock \showarticletitle{{Z}ipporah: a Fast and Scalable Data Cleaning
  System for Noisy Web-Crawled Parallel Corpora}. In
  \bibinfo{booktitle}{\emph{EMNLP}}.
\newblock


\bibitem[\protect\citeauthoryear{Zhao, Dua, and Singh}{Zhao
  et~al\mbox{.}}{2018}]%
        {zhao2017generating}
\bibfield{author}{\bibinfo{person}{Zhengli Zhao}, \bibinfo{person}{Dheeru Dua},
  {and} \bibinfo{person}{Sameer Singh}.} \bibinfo{year}{2018}\natexlab{}.
\newblock \showarticletitle{Generating natural adversarial examples}. In
  \bibinfo{booktitle}{\emph{ICLR}}.
\newblock


\end{thebibliography}

\end{document}